\documentclass[letterpaper]{article} 
\usepackage{aaai23}  
\usepackage{times}  
\usepackage{helvet}  
\usepackage{courier}  
\usepackage[hyphens]{url}  
\usepackage{graphicx} 
\usepackage{natbib}  
\usepackage{caption} 
\usepackage{algorithm}
\usepackage{algorithmic}

\usepackage{xcolor}         
\usepackage{hyperref}
\usepackage{navigator}
\usepackage{subfigure}
\usepackage{multirow}
\usepackage{bbm}
\usepackage{adjustbox}
\usepackage{bbding}
\usepackage{ntheorem}
\usepackage{enumitem}
\usepackage{oplotsymbl}
\usepackage{amsfonts}

\newlength\savewidth\newcommand\shline{\noalign{\global\savewidth\arrayrulewidth
		\global\arrayrulewidth 1pt}\hline\noalign{\global\arrayrulewidth\savewidth}}

\usepackage{newfloat}
\usepackage{listings}
\DeclareCaptionStyle{ruled}{labelfont=normalfont,labelsep=colon,strut=off} 
\floatstyle{ruled}
\newfloat{listing}{tb}{lst}{}
\floatname{listing}{Listing}

\definecolor{mydarkblue}{rgb}{0,0.08,0.45}
\hypersetup{
	colorlinks=true,
	urlcolor=magenta,
	citecolor=mydarkblue,
}

\title{DQ-DETR: Dual Query Detection Transformer for \\Phrase Extraction and Grounding}
\author {
    Shilong Liu\textsuperscript{\rm 1,2}\thanks{This work was done when Shilong Liu, Yaoyuan Liang, Feng Li, Shijia Huang, and Hao Zhang were interns at IDEA.},
    Yaoyuan Liang\textsuperscript{\rm 3},
    Feng Li\textsuperscript{\rm 2,4},
    Shijia Huang\textsuperscript{\rm 5},
    Hao Zhang\textsuperscript{\rm 2,4},\\
    Hang Su\textsuperscript{\rm 1},
    Jun Zhu\textsuperscript{\rm 1\dag},
    Lei Zhang\textsuperscript{\rm 2}\thanks{Corresponding author.}
}
\affiliations {
    \textsuperscript{\rm 1} Dept. of Comp. Sci. and Tech., BNRist Center, State Key Lab for Intell. Tech. \& Sys., Institute for AI, Tsinghua-Bosch Joint Center for ML, Tsinghua University. \quad
    \textsuperscript{\rm 2} International Digital Economy Academy (IDEA).\\
    \textsuperscript{\rm 3} Tsinghua-Berkeley Shenzhen Institute, Tsinghua University. \quad
    \textsuperscript{\rm 4} The Hong Kong University of Science and Technology. \\
    \textsuperscript{\rm 5} The Chinese University of Hong Kong. \quad

{\small{\{liusl20,liang-yy21\}@mails.tsinghua.edu.cn}},
{\small{\{fliay,hzhangcx\}@connect.ust.hk}},
{\small{sjhuang@cse.cuhk.edu.hk}},
{\small{\{suhangss,dcszj\@mail.tsinghua.edu.cn}},
{\small{\{leizhang\}@idea.edu.cn}}
}

\usepackage{bibentry}

\begin{document}

\maketitle

\begin{abstract}
In this paper, we study the problem of visual grounding by considering both phrase extraction and grounding (PEG). In contrast to the previous phrase-known-at-test setting, PEG requires a model to extract phrases from text and locate objects from image simultaneously, which is a more practical setting in real applications. As phrase extraction can be regarded as a $1$D text segmentation problem, we formulate PEG as a dual detection problem and propose a novel DQ-DETR model, which introduces dual queries to probe different features from image and text for object prediction and phrase mask prediction. Each pair of dual queries is designed to have shared positional parts but different content parts. Such a design effectively alleviates the difficulty of modality alignment between image and text (in contrast to a single query design) and empowers Transformer decoder to leverage phrase mask-guided attention to improve the performance. To evaluate the performance of PEG, we also propose a new metric CMAP (cross-modal average precision), analogous to the AP metric in object detection. The new metric overcomes the ambiguity of Recall@1 in many-box-to-one-phrase cases in phrase grounding. As a result, our PEG pre-trained DQ-DETR establishes new state-of-the-art results on all visual grounding benchmarks with a ResNet-101 backbone. For example, it achieves $91.04\%$ and $83.51\%$ in terms of recall rate on RefCOCO testA and testB with a ResNet-101 backbone. 
Code will be availabl at \url{https://github.com/IDEA-Research/DQ-DETR}.
 
\end{abstract}

\section{Introduction}
\label{sec:intro}
\begin{figure}[!t]
\centering
\vspace{-0.3cm}
\resizebox{1.0\columnwidth}{!}{%
\includegraphics{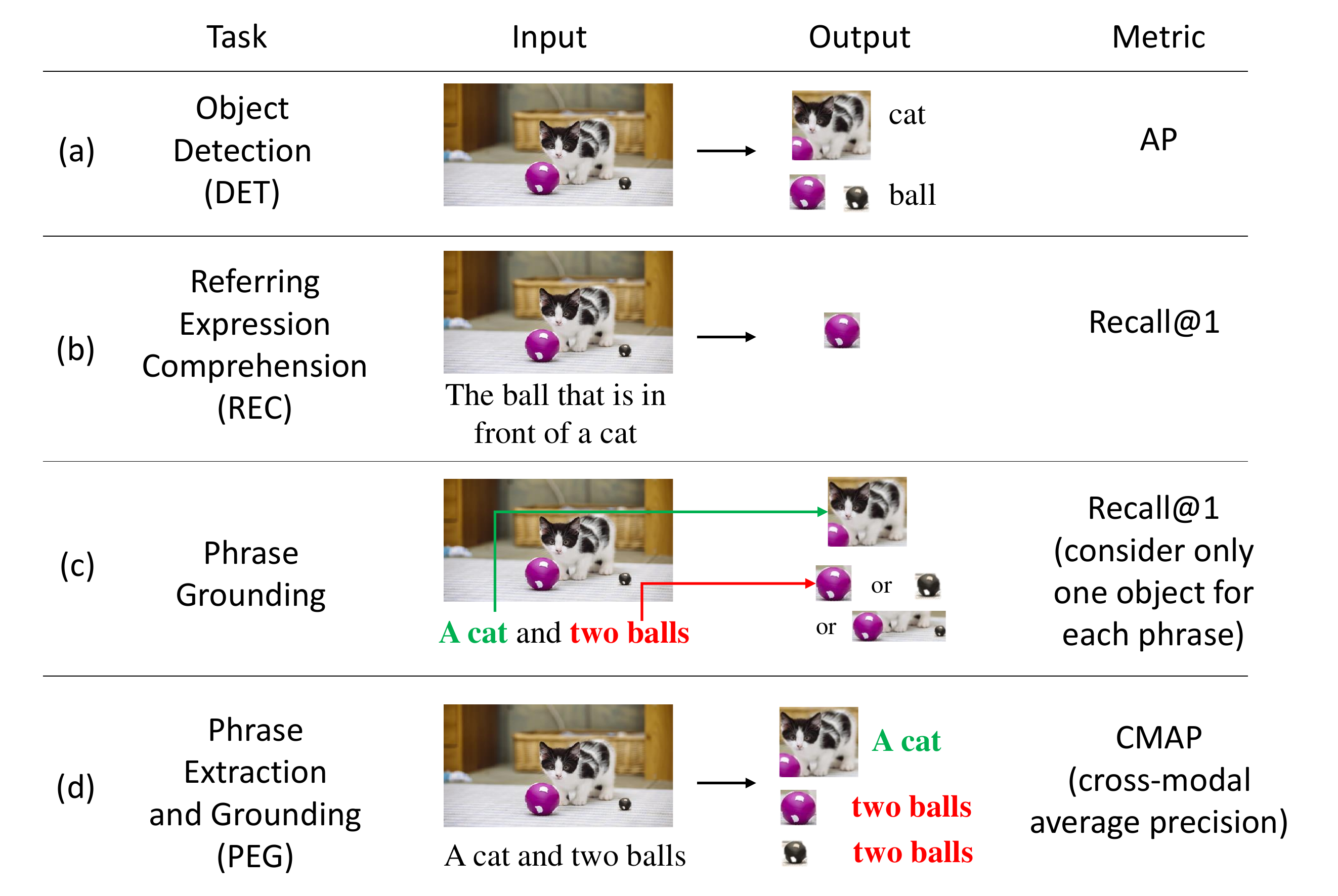}}
\vspace{-0.4cm}
\caption{
\small{
Comparisons of different settings. 
(a) Given an image, object detection (DET) is to locate objects in pre-defined categories. The most popular metric for DET is mAP \cite{lin2015microsoft}. 
(b) Referring expression comprehension (REC) aims at locating objects described by an input text. Its performance is normally evaluated by the recall of the most confident outputs.
(c) Phrase grounding needs to ground the spatial regions described by phrases in an image. Most methods treat this task as a ranking problem and evaluate it by recall. The phrases in sentences are usually assumed known during inference.
(d) We re-emphasize a phrase extraction and grounding (PEG) setting in this paper. A key difference compared with phrase grounding is that phrases in PEG are unknown during test. We propose a CMAP (cross-modal average precision) metric for the PEG task, analogous to mAP for DET.
In this paper, we use the term ``visual grounding" for all of three tasks: REC, phrase grounding, and PEG.
}}
\label{fig:task_comparisons}
\end{figure}

Visual grounding aims to locate objects referred to by language expressions or phrases, which closely relates to object detection (DET, Fig. \ref{fig:task_comparisons} (a)) in vision. It has received increasing attention for its potential to benefit other multi-modal tasks like visual question answering (VQA)~\cite{AkiraFukui2016MultimodalCB} and image retrieval~\cite{AndrejKarpathy2014DeepFE, FilipRadenovic2016CNNIR}.

Some works~\cite{JiajunDeng2021TransVGEV, JianqiangHuang2022DeconfoundedVG} treat the terms \textit{visual grounding}, \textit{referring expression comprehension (REC)}, and \textit{phrase grounding} interchangeable. However, they have subtle differences. Both REC and phrase grounding are sub-tasks of visual grounding.

REC locates objects with a free-form guided text, as shown in Fig. \ref{fig:task_comparisons} (b). It has only one category of objects to detect as requested by a referring expression, while phrase grounding needs to find all objects mentioned in a caption, as shown in Fig. \ref{fig:task_comparisons} (c). Though they have different definitions, phrase grounding can be reformulated as a REC task by extracting phrases as referring expressions since phrases are assumed known during test~\cite{JiajunDeng2021TransVGEV, YeDu2021VisualGW, JianqiangHuang2022DeconfoundedVG}. 
Some methods~\cite{ZongshenMu2021DisentangledMG, YongfeiLiu2019LearningCC} use non-REC solutions for phrase grounding, while they also treat phrases as known during test.

We argue it is more practical to treat phrases as unknown during test and study the problem of visual grounding in this paper by considering both phrase extraction and grounding (PEG), as shown in Fig. \ref{fig:task_comparisons} (d). 
Solving PEG by developing a large-scale image-text-paired training dataset with both phrases and objects annotated is prohibitively costly.
A simple way to extend existing REC models~\cite{PengWang2022UNIFYINGAT,ChaoyangZhu2022SeqTRAS} to PEG is to develop a two-stage solution: firstly extracting phrases using an NLP tool like spaCy~\cite{spacy2} and then applying a REC model. However, such a solution may result in inferior performance (as shown in our Table \ref{tab:pretrain}) as there is no interaction between the two stages. For example, an image may have no object or more than one object that corresponds to an extracted phrase. Yet most REC models~\cite{Miao2022ReferringEC,ChaoyangZhu2022SeqTRAS} predict only one object for each extracted phrase. 
Let alone inaccurate phrase extraction can mislead a REC model to predict unrelated objects.

We are not the first to propose the PEG setting. Some previous works~\cite{AndrejKarpathy2014DeepFE, AndrejKarpathy2014DeepVA} align image regions and phrases for image retrieval. Flickr30k Entities~\cite{BryanAPlummer2015Flickr30kEC} evaluates models under the scenario in which phrases are unknown as well. They extract noun phrases using NLP tools and penalize recall if the tools extract inaccurate phrases. Despite of such early explorations, most successors~\cite{MohitBajaj2019G3raphGroundGL,JiajunDeng2021TransVGEV} treat phrase grounding as a retrieval task and use ground truth phrases as inputs. Hence we re-emphasize the PEG setting, where we predict object-phrase pairs given only a pair of image and text as input, without assuming phrases as known input. We can reformulate all three other tasks (DET, REC, and phrase grounding) as PEG tasks. 

\begin{figure}[!t]
\centering
\vspace{-0.3cm}
\resizebox{1.0\columnwidth}{!}{%
\includegraphics{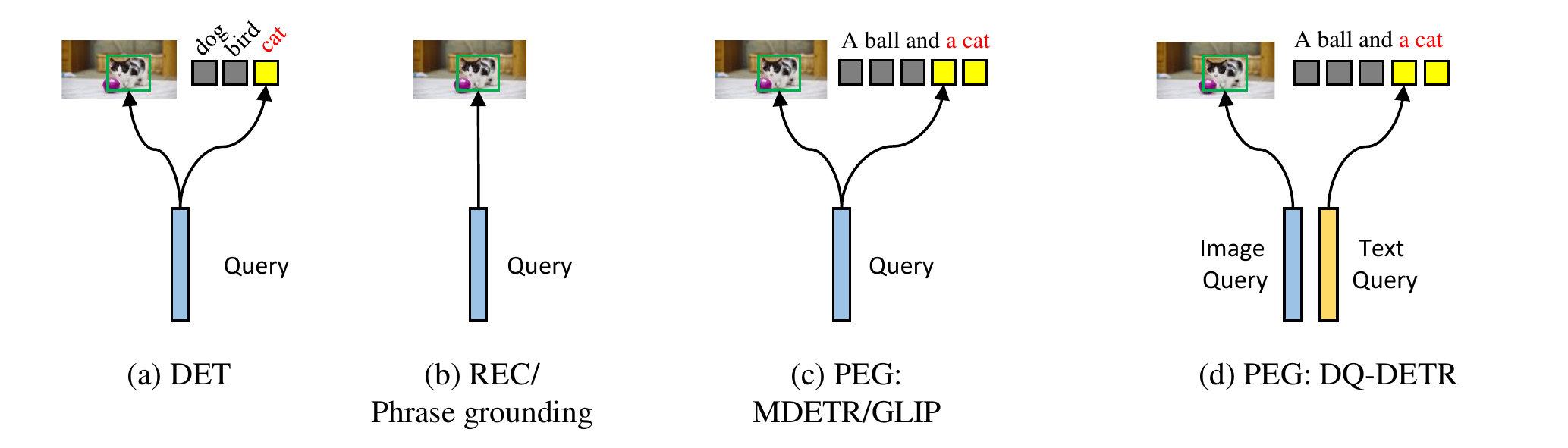}}
\vspace{-0.6cm}
\caption{
\small{
Comparisons of different models. The ``query" here refers to the input of Transformer decoder.
(a) Each query in a DET model corresponds to an object and a class label id. 
(b) Each query in REC/phrase grounding models predicts an object only.
(c) Some previous work suitable for PEG, such as MDETR~\cite{kamath2021mdetr} and GLIP~\cite{li2021grounded}, performs box localization and phrase extraction using the same queries.
(d) We propose to decouple a single query into dual queries in our DQ-DETR for two tasks: object localization and phrase extraction, which are for two different modalities.
}}
\label{fig:dual_branch}
\end{figure}

PEG lifts the importance of phrase extraction, which is often overlooked in previous task formulations and solutions. Some previous works can be used for the PEG task, such as MDETR~\cite{kamath2021mdetr} and GLIP~\cite{li2021grounded}, which use the same query (in a DETR framework) for both object localization and phrase extraction, as shown in Fig. \ref{fig:dual_branch} (c). However, their phrase extraction module requires a query to have an extra capability to perform the challenging image-text feature alignment, which can interfere with the bounding box regression branch and result in an inferior performance (See Sec. \ref{sec:why} for a more detailed discussion). 

We note that phrase extraction is to localize a noun phrase from an input text, which can be regarded as a 1D text segmentation problem that predicts a 1D text mask for a target phrase. Such a problem is analogous to 2D mask prediction for an object instance in 2D image segmentation. Especially, inspired by recent progress of DETR-like models (e.g., DINO~\citep{zhang2022dino},  Mask2Former~\citep{BowenCheng2022Mask2FormerFV}), we develop a more principled solution DQ-DETR, which is a dual query-based\footnote{We use the term ``query'' as the input of the Transformer decoder layers in this paper, following the common practice in the Transformer and DETR-related literature~\citep{vaswani2017attention, meng2021conditional, liu2022dabdetr}. The definition differs from some visual grounding papers, where ``query" refers to an input text. A detailed explanation of our dual query is available in the appendix.}  
DETR-like model for PEG. As shown in Fig.~\ref{fig:dual_branch} (d), our model uses dual queries to perform object detection and text mask prediction in one DETR framework. The text mask prediction is very similar to instance mask prediction as in Mask2Former, hence we can use masked-attention Transformer decoder layers to improve the performance of text mask prediction. In DQ-DETR, a pair of dual queries is designed to have shared positional parts but different content parts\footnote{A DETR query consists of two parts: a content part and a positional part. More detailed discussions can be referred to \cite{meng2021conditional} and \cite{liu2022dabdetr}.}. Such a decoupled query design helps alleviate the difficulty of modality alignment between image and text, yielding faster convergence and better performance. 

\begin{figure}[!t]
\centering
\vspace{-0.3cm}
\resizebox{1.0\columnwidth}{!}{%
\includegraphics{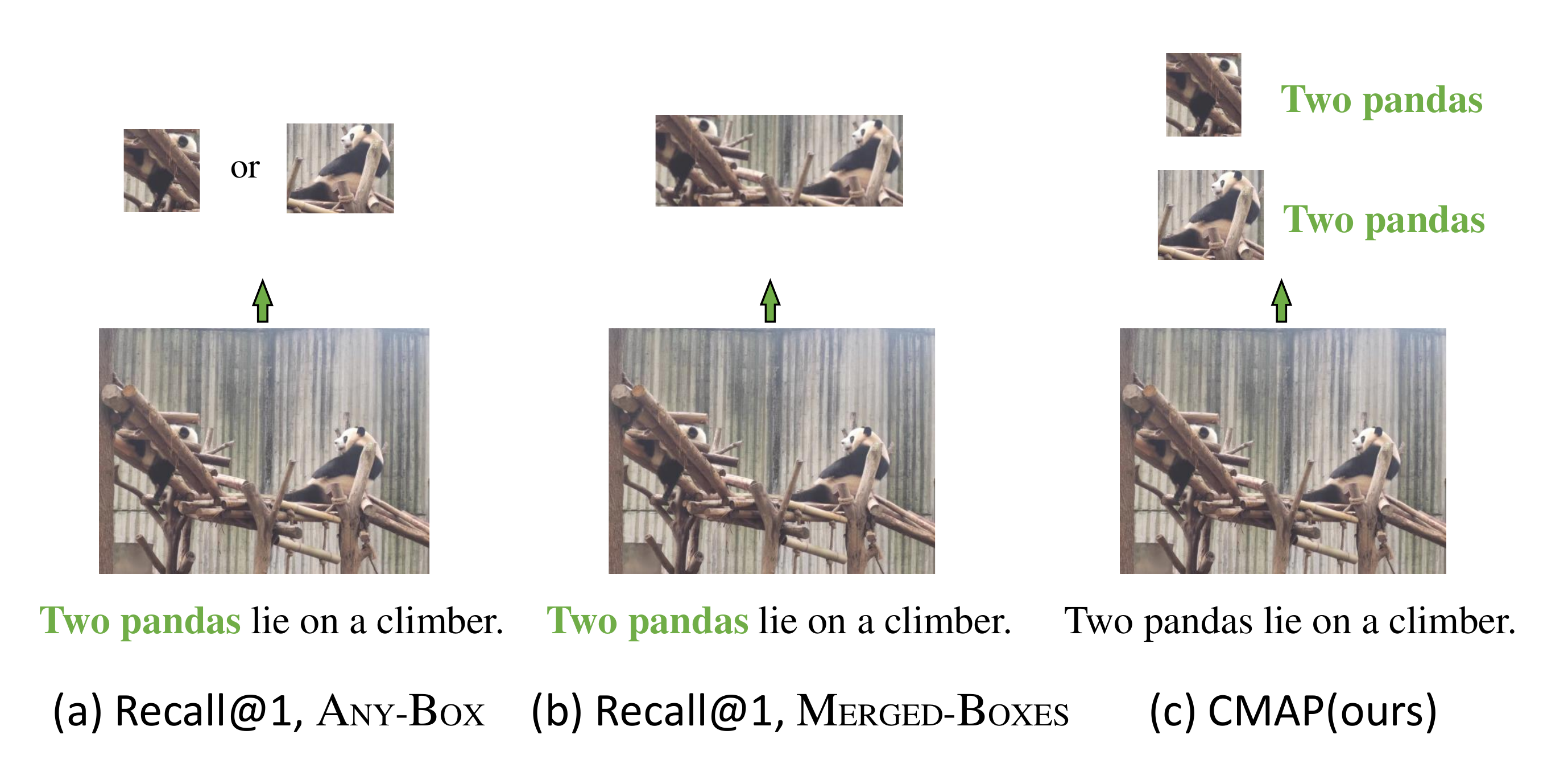}}
\vspace{-0.8cm}
\caption{
\small{
Comparisons of different metrics. We only plot objects corresponding to the phrases ``Two pandas'' for a better comparison. (a) and (b) are used for phrase grounding. 
(a) The \textsc{Any-Box} setting treats a prediction as correct if any of the ground truth boxes is matched.
(b) \textsc{Merged-Boxes} combines all objects for one phrase to a big box for evaluation.
(c) Our metric CMAP encourages a model to predict all objects and their corresponding phrases.
}}
\label{fig:cmap}
\end{figure}

To evaluate models on the PEG setting, we propose a new metric CMAP ({cross-modal average precision}), which is analogous to the AP metric widely used in object detection. It measures the accuracy of both phrase extraction and object localization, as shown in Fig. \ref{fig:cmap} (c). The CMAP metric overcomes the ambiguous issue of the previous Recall@1 when multiple objects correspond to one phrase. Recall@1 evaluates the accuracy of boxes with the highest confidence. However, for cases where multiple objects correspond to one phrase, the metric becomes ambiguous. To deal with such cases, previous works~\cite{BryanAPlummer2015Flickr30kEC, LiunianHaroldLi2019VisualBERTAS, kamath2021mdetr} leveraged two different protocols, which we denote as \textsc{Any-Box} (Fig. \ref{fig:cmap} (a)) and \textsc{Merged-Boxes} (Fig. \ref{fig:cmap} (b)) protocols following MDETR~\cite{kamath2021mdetr}. The \textsc{Any-Box} setting treats a prediction as correct if any of the ground truth boxes is matched. However, it cannot evaluate a model's capability of finding all objects in an image. The other protocol, \textsc{Merged-Boxes}, combines all objects for one phrase into a big box for evaluation. While being able to capture all objects, this protocol cannot measure the localization accuracy for every object instance.

We summarize our contributions as follows:
\begin{enumerate}
    \item By comparing three settings in visual grounding: DET, REC, and phrase grounding, we re-emphasize a PEG setting, which is often overlooked in previous works. 
    To take the phrase extraction accuracy into account, we propose a new cross-modal average precision (CMAP) metric for PEG to measure a combined accuracy for both phrase extraction and object localization. The CMAP metric is free of confusion when multiple objects correspond to one phrase.
    \item We interpret noun phrase extraction as a 1D text segmentation problem and formulate the PEG problem as predicting both bounding boxes for objects and text masks for phrases. Accordingly, we develop a novel dual query-based DETR-like model DQ-DETR with several techniques to improve the performance of phrase extraction and object localization.
    \item We validate our methods on several benchmarks and establish new state-of-the-art results, including Flickr30k, RefCOCO/+/g, and COCO. Our model obtains $76.0\%$ CMAP$_{50}$ and $83.2\%$ Recall@1 at Flickr30k entities~\cite{BryanAPlummer2015Flickr30kEC}. Moreover, we achieve $91.04\%$ and $83.51\%$ in terms of recall rate on RefCOCO testA and testB with a ResNet-101 backbone.
\end{enumerate}

\section{PEG \& CMAP}
We present the PEG (phrase extraction and grounding) problem formulation and the CMAP (cross-modal average precision) definition in this section. 

Given an image-text pair as input, PEG requires a model to predict region-phrase pairs from the input image and text pair, as shown in Fig. \ref{fig:cmap}. The PEG task can be viewed as a dual detection problem for image box detection and text mask segmentation, since noun phrase extraction can be interpreted as a 1D text segmentation problem. 

To measure both the text phrase extraction accuracy and the image object localization accuracy, we propose a new metric which is similar to the AP metric used in DET. AP is calculated by integrating the area under a P-R curve. The key to plot P-R curves is to decide positive and negative samples. DET benchmarks like COCO~\cite{lin2015microsoft} leverage IOU (intersection over union) between a predicted box and a ground truth box to discriminate positive and negative predictions. As we interpret phrase extraction as a $1$D segmentation problem, we use dual IOU to choose positive predictions. The dual IOU is defined as:

\begin{equation}
    \mathrm{IOU}_{\mathrm{dual}} = (\mathrm{IOU}_{\mathrm{box}})^{0.5} \times \mathrm{IOU}_{\mathrm{phrase}},
\end{equation}
where $\mathrm{IOU}_{\mathrm{box}}$ is the box IOU and $\mathrm{IOU}_{\mathrm{phrase}}$ is the phrase IOU. We take the square root of $\mathrm{IOU}_{\mathrm{box}}$ to make $\mathrm{IOU}_{\mathrm{dual}}$ a two dimensional metric so that its threshold (e.g. 0.5) has a similar meaning to $\mathrm{IOU}_{\mathrm{box}}$. Following the common practice in phrase grounding and REC, we use $\mathrm{IOU}_{\mathrm{dual}} >= 0.5$ as positive samples, and vice versa. We use the term ``CMAP$_{50}$'' to denote the metric at threshold $0.5$.

\section{DQ-DETR}
\begin{figure}[t]
 \centering
 \vspace{-0.3cm}
 \resizebox{\columnwidth}{!}{%
 \includegraphics{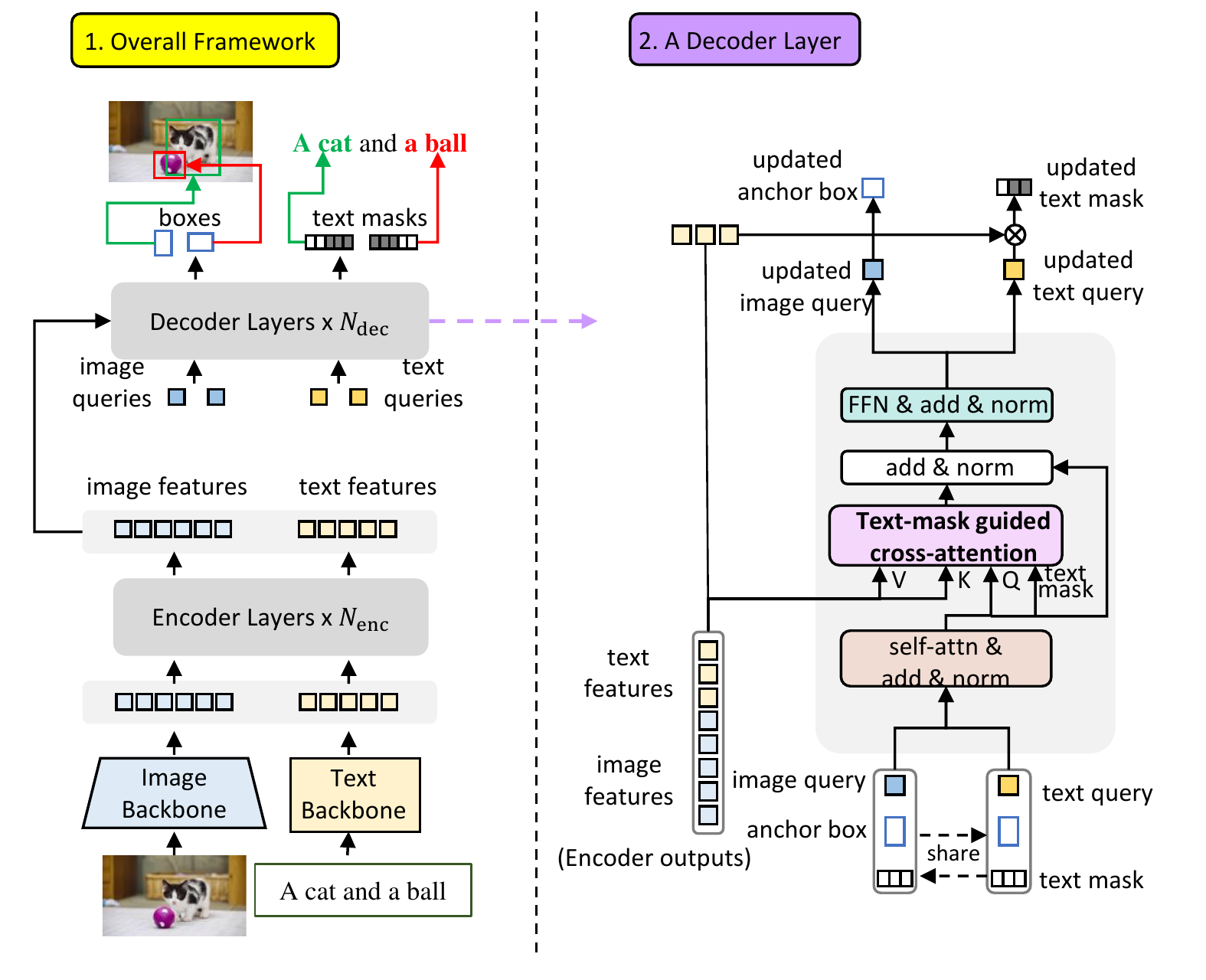}}
 \vspace{-0.9cm}
 \caption{\small{The framework of our proposed DQ-DETR model. The left block is the overall framework. The right block presents the detailed structure of a decoder layer in DQ-DETR.}}
 \vspace{-0.2cm}
 \label{fig:framework}
 \end{figure}

Following DETR~\citep{carion2020end} and MDETR~\citep{kamath2021mdetr}, DQ-DETR is a Transformer-based encoder-decoder architecture, which contains an image backbone, a text backbone, a multi-layer Transformer encoder, a multi-layer Transformer decoder, and several prediction heads. 

Given a pair of inputs \texttt{(Image, Text)}, we extract image features and text features using an image backbone and a text backbone, respectively. The image and text features are flattened, concatenated, and then fed into the Transformer encoder layers. 
We then use learnable dual queries for the decoder layers to probe desired features from the concatenated multi-modality features. The image queries and text queries will be used for box regressions and phrase localizations, respectively, as shown in Fig. \ref{fig:framework} left.

\subsection{Why Decoupling Queries for Image Box Prediction and Text Mask Prediction?}
\label{sec:why}

\begin{figure*}[th]
\centering
\subfigure[ ]{
\begin{minipage}[t]{0.33\linewidth}
\centering
\vspace{-0.1cm}
\includegraphics[width=\linewidth]{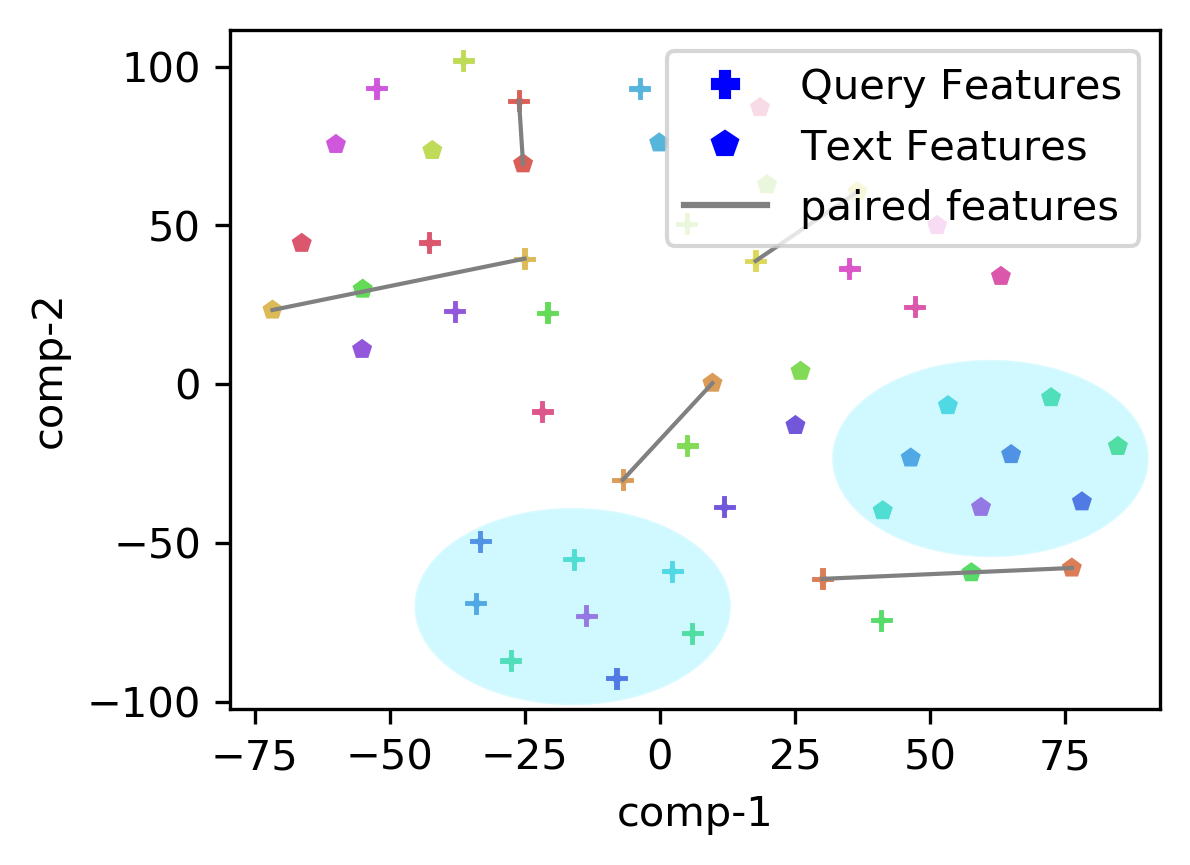}
\vspace{-0.4cm}
\label{fig:mdetr_feature_tsne}
\end{minipage}
}%
\subfigure[ ]{
\begin{minipage}[t]{0.32\linewidth}
\centering
\vspace{-0.1cm}
\includegraphics[width=\linewidth]{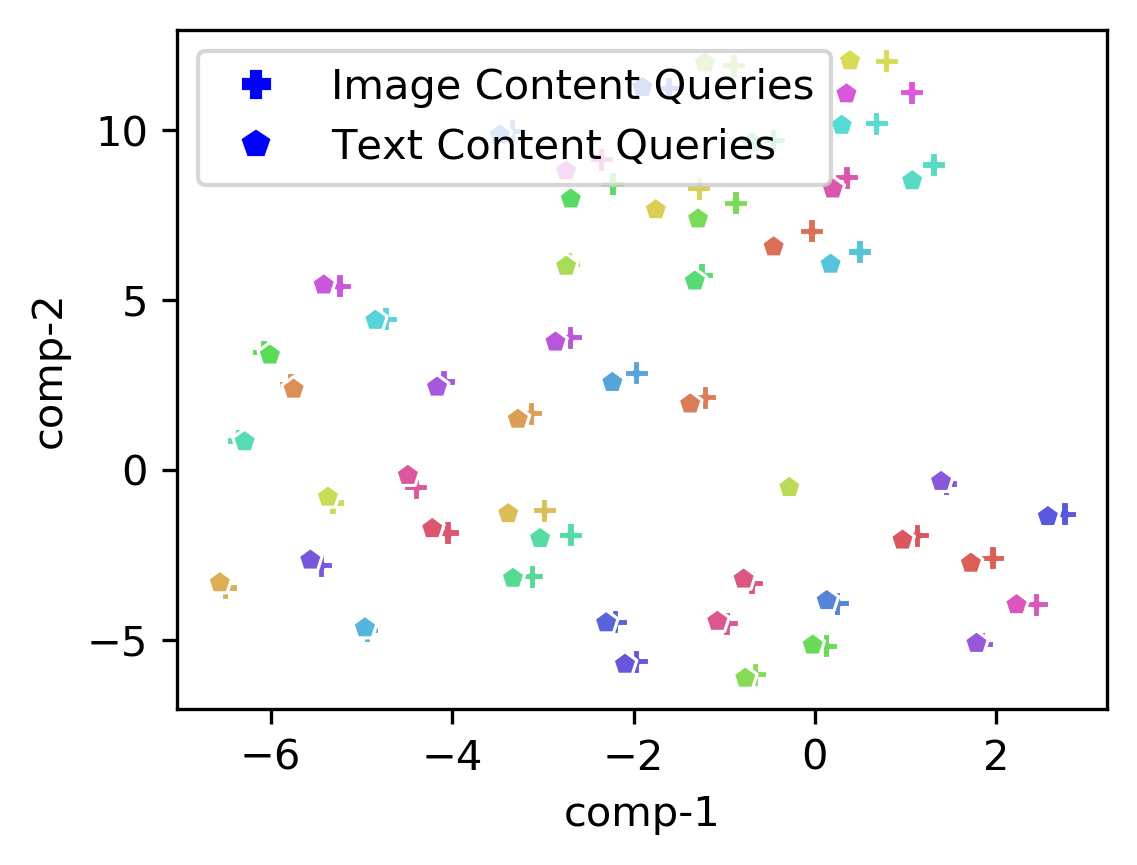}
\vspace{-0.4cm}
\label{fig:dmdetr_query_pair}
\end{minipage}
}%
\subfigure[ ]{
\begin{minipage}[t]{0.33\linewidth}
\centering
\vspace{-0.1cm}
\includegraphics[width=\linewidth]{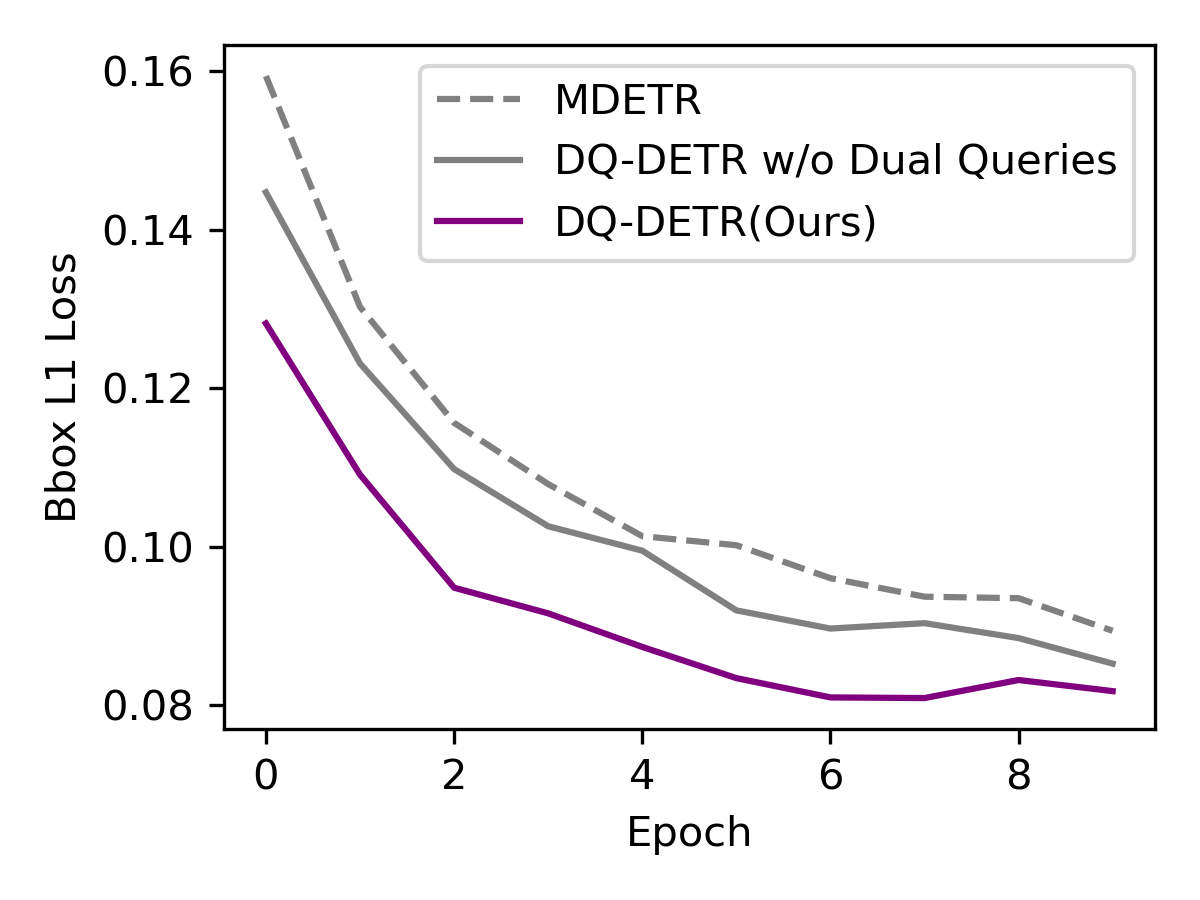}
\vspace{-0.4cm}
\label{fig:compare_bbox_l1}
\end{minipage}
}%
\centering
\vspace{-0.4cm}
\caption{\small{(a) Visualization of query and text features in MDETR~\citep{kamath2021mdetr} by t-SNE. Different colors are for different pairs of features. Some paired features are linked by gray lines for easy check. The ``comp-1'' and ``comp-2'' are the main principal components after dimensionality reduction. Some clustering features are shaded in blue. (b) Visualization of dual queries in our DQ-DETR by t-SNE. Different colors are for different pairs of dual queries. (c) Comparison of test bounding box L1 loss vs. training epochs. (Best viewed in color.) }
}
\end{figure*}

Most object detectors~\citep{carion2020end,zhu2020deformable,liu2022dabdetr} perform box regression and label prediction with the same queries, as shown in Fig. \ref{fig:dual_branch} (a). 
For phrase grounding, a query in previous models not only needs to regress a bounding box of a target object but also needs to localize its corresponding noun phrase in a query text. 
Such a one-query design drives them to design a contrastive learning way to align multi-modality features. However, modality alignment is a very challenging task. We take MDETR as an example and visualize the paired features in Fig. \ref{fig:mdetr_feature_tsne}. MDETR uses a contrastive alignment loss to encourage modality alignment between queries and text features, where it assumes queries contain image features. 
We use the MDETR model that is pre-trained on $1.3$M data provided by the official MDETR GitHub page\footnote{The open-sourced code of MDETR we used: \url{https://github.com/ashkamath/mdetr}.} for visualization.
Each paired features have the same colors in the figure. We also link some pairs of features with gray lines for a better visualization. The results show that most paired features in MDETR cannot be aligned well. For example, some paired features are farther away. Moreover, some features of the same modality tend to cluster with each other rather than cluster with their paired features of the other modalities, as shown in the blue-shaded regions.

The alignment challenge across two modalities is not an isolated phenomenon. It has been widely recognized on even larger-scale pre-trained models like CLIP~\citep{AlecRadford2021LearningTV, AndreasFrst2021CLOOBMH}. 
CLIP-like models aim to align multi-modality features only, whereas MDETR/GLIP also needs to regress bounding boxes. 
We speculate that the two prediction tasks, box prediction and phrase localization, need different features to accommodate the modality gap.
Hence we design a dual query model for visual grounding with different queries for different modalities. 
To confirm our speculation, we visualize the dual queries at the Transformer decoder output in Fig. \ref{fig:dmdetr_query_pair}. It shows that each paired dual features (a ``$\pentagofill$''-shape point and a ``$\textbf{+}$''-shape point with the same color) have similar but different features.

As the desired features for bounding box prediction and phrase localization are different, we suspect that the alignment between image features and text features can interfere with the bounding box regression branch. To verify this, we compare the L1 box losses of three models, MDETR, DQ-DETR without dual queries, and DQ-DETR (with dual queries) , in Fig. \ref{fig:compare_bbox_l1}. All models are trained with ResNet-50~\citep{he2015deep} backbones on the Flickr30k Entities dataset~\citep{BryanAPlummer2015Flickr30kEC} for ten epochs. 
It shows that decoupling queries helps the training of box prediction branches and can accelerate the convergence of the box regression module.

\subsection{Dual Queries for Dual Detections}
\label{sec:dual_query}
We propose to decouple the queries for bounding box regression and phrase localization in DQ-DETR. 
However, as the dual queries aim to predict paired \texttt{(Region, Phrase)} results, both queries need to focus on the same region of an object in the input image and the same position of a phrase in the input text.
Hence we propose to share the positional parts and decouple the content parts of the queries. 
As we formulate the problem as a dual detection problem for image box detection and text phrase segmentation, we introduce two items for the positional queries, i.e., image positional queries and text positional queries.
More concretely, the image positional queries are formulated as anchor boxes like DAB-DETR~\citep{liu2022dabdetr} and then projected to high dimensions with sine/cosine encoding. The text positional queries are formulated as 1D segmentation masks like Mask2Former~\citep{BowenCheng2022Mask2FormerFV} and then used for text mask-guided attention. 
The image positional queries are predicted by the updated image queries, and the text positional queries are generated by performing dot product between the updated image queries and text features from the encoder output, as shown in Fig. \ref{fig:framework} right. Both two positional queries will be shared by the dual queries as positional parts for the next layer.
Beyond the positional and content parts, we add a learnable modality embedding to the features of different modalities.
We list the components of our queries in Table \ref{tab:queries}.  

\begin{table}[h]
    \centering
        \footnotesize
            \renewcommand{\arraystretch}{1.3}
    \resizebox{0.96\columnwidth}{!}{%
    \begin{tabular}{lll}
        \shline
        {Queries} & Image Queries $Q_{I} \in \mathbb{R}^{N_q,D}$ & Text Queries $Q_{T}\in\mathbb{R}^{N_q,D}$ \\
        \hline
        {Content Parts} & $Q^{(C)}_{I} \in \mathbb{R}^{N_q,D}$ & $Q^{(C)}_{T} \in \mathbb{R}^{N_q,D}$ \\
        {Image Positional Parts} & $A_{I} \in \mathbb{R}^{N_q,4}$ &  $A_{T}=A_{I}$ \\
        {Text Positional Parts}  & $M_{I}=M_{T}$ & $M_{T} \in {\{0,1\}}^{N_{\mathrm{text}}}$ \\
        {Modality Embeddings} & $\mathrm{ModalityToken}_{I} \in \mathbb{R}^{D}$ & $\mathrm{ModalityToken}_{T} \in \mathbb{R}^{D}$ \\
        \shline
    \end{tabular}}
    \vspace{-0.2cm}
    \caption{
    \small{
    The table of components of our dual queries. We use the notations $N_q$ for the number of paired queries, $N_{\mathrm{text}}$ for the length of text tokens, and $D$ for feature dimension.
    We formulate image positional queries $A_{I}$ as anchor boxes and text positional queries $M_{T}$ as 1D binary masks with the same length as the text tokens. 
    Both image and text queries have shared positional parts, including image positional parts and text positional parts, but different content parts. 
    }
    }
    \label{tab:queries}
\end{table}

\subsection{Text Mask-Guided Attention}
\label{sec:text_attn}
The 1D segmentation formulation of phrase localization inspires us to
propose a text mask-guided attention to let queries focus on phrase tokens of interest, analogous to the mask attention in Mask2Former~\citep{BowenCheng2022Mask2FormerFV}. 
Each text query has a text positional query $M_T\in \mathbbm{1}^{N_{\mathrm{text}}}$, which is a binary mask with the same length as the text features. 
We use the encoder output image-text-concatenated features as keys and values for cross-attention.
The binary masks will be used as attention masks for the text features in the concatenated features. 
Text features will be used if their corresponding mask values are ones, otherwise they will be masked out. This operation constrains the attention on the target phrases while the predicted masks are updated layer by layer to to get closer to ground truth masks.

We use all-ones masks as inputs for the first decoder layer. Other layers will leverage the predicted masks from their previous layers. 
The final updated masks are the outputs for phrase localization.

\subsection{Loss Functions}
\label{sec:loss}
Following DETR~\citep{carion2020end} and MDETR~\citep{kamath2021mdetr}, we use bipartite matching to assign ground truth object boxes and text phrases to dual queries during training. 
The final loss functions can be grouped into boxes losses for images and phrase losses for texts.
We use the L1 loss and the GIOU~\citep{rezatofighi2019generalized} loss for bounding box regression. 
For phrase localization, we use a contrastive softmax loss. 

For a text query at the output of the decoder $Q^{\mathrm{(out)}}\in \mathbb{R}^{N_q, D}$, we compute the similarities between this query and the encoder output text features $F^{\mathrm{(enc)}}_T\in \mathbb{R}^{N_{\mathrm{text}}, D}$ to predict a text segmentation mask.
We first linearly project the decoder output $Q^{\mathrm{(out)}}$ to get  $Q=\mathrm{Linear}_Q(Q^{\mathrm{(out)}}) \in \mathbb{R}^{N_q, D_1}$. Then we linearly project the encoder output text feature $F^{\mathrm{(enc)}}_T$ to get $F_T= \mathrm{Linear}_T(F^{\mathrm{(enc)}}_T) \in \mathbb{R}^{N_{\mathrm{text}}, D_1}$. The notations $D_1$ is the dimension of the projected space, and $\mathrm{Linear}_Q, \mathrm{Linear}_T$ are two linear layers. 
As some queries may not correspond to any desired phrase, similar to a DETR query not matching with any ground truth object, we set an extra learnable \texttt{no\_phrase} token $\mathrm{NoPhraseToken} \in \mathbb{R}^{D_1}$ for \texttt{no\_phrase} queries. We then concatenate the projected text feature and the \texttt{no\_phrase} token to get an extended text feature $F_T' = \mathrm{Concat}(F_T, \mathrm{NoPhraseToken}) \in \mathbb{R}^{N_{\mathrm{text}}+1, D_1}$. 

The final contrastive softmax loss is performed between the projected query features $Q \in \mathbb{R}^{N_q, D_1}$ and the extended text feature $F_T' \in \mathbb{R}^{N_{\mathrm{text}}+1, D_1}$. Let $S_{q_i}$ be the set of text token indices of a target phrase for a given query $q_i \in Q$. The phrase localization loss for query $q_i$ is:

\begin{equation}
\label{eq:contrastive_softmax}
\mathcal{L}_{\mathrm{phrase},i} =  
    \sum_{j \in S_{q_i}}
        (-\mathrm{log} \frac{q_i^\top p_j/\tau}{\sum_{k=0}^{N_{\mathrm{text}}+1} q_i^\top p_k/\tau})
, 
\end{equation}
where $\tau$ is a temperature parameter which is empirically set to $0.07$ in our experiments, and $p_j \in F_T'$ is a text feature or a \texttt{no\_phrase} token with index $j$. We down-weight the loss by $0.05$ when no objects are assigned to the query $q_i$ to balance the classes.

\section{Experiments}
\subsection{Implementation Details}

\textit{Models.} 
We use two commonly used image backbones, ResNet-50 and ResNet-101~\citep{he2015deep} pre-trained on  ImageNet~\citep{deng2009imagenet}, for our base setting and pre-training setting, respectively. Both two pre-trained models are provided by PyTorch~\citep{paszke2017automatic}.
For the text backbones, we use the pre-trained RoBERTa-base~\citep{YinhanLiu2019RoBERTaAR} provided by HuggingFace~\citep{wolf2019huggingface} in our experiments. 
We set $D=256$ and $D_1=64$ in our implementations and use $100$ pairs of dual queries.
Our models use $6$ encoder layers and $6$ decoder layers. 
The learning schedules are different for different settings, which will be described in each subsection.
The initial learning rates for the Transformer encoder-decoder and image backbone are $1e^{-4}$ and $1e^{-5}$, respectively.
For the text backbone, we use a linear decay schedule from $5e^{-5}$ to $0$ with a linear warm-up in the first $1\%$ steps.
To stabilize the bipartite graph matching, we use anchor denoising~\citep{li2022dn} in our implementations.

\subsection{The Pre-training Setting}
\label{sec:pretrain_setting}

\subsubsection{4.2.1 Pre-training Task: PEG}
\label{sec:pre_train}
Following MDETR~\citep{kamath2021mdetr}, we use the combined dataset of Flickr30k, COCO, and Visual Genome for our pre-training. The backbone we used is ResNet-101.
We pre-train our model on the combined dataset for $25$ epochs and drop the initial learning rate by $10$ after the $20$-th epoch. The pre-training takes about $100$ hours on $16$ Nvidia A100 GPUs with $4$ images per GPU. We then fine-tune the model on different tasks with $4$ GPUs, except for the object detection task, which needs $8$ GPUs.

We compare our model with three baselines in Table \ref{tab:pretrain}. 
We use the state-of-the-art REC model OFA-REC~\cite{PengWang2022UNIFYINGAT}
\footnote{We use the $\mathrm{OFA}_{\mathrm{Base}}$ provided in \url{https://github.com/OFA-Sys/OFA}.}
to demonstrate the necessary of our PEG. OFA is an unified model pre-trained with more than 50M images and can be used for REC tasks. To adapt it to our PEG task, we use spaCy~\cite{spacy2} to extract noun phrases. The results show that OFA-REC+spaCy is much worse than the other two end-to-end models in terms of the CMAP$_{50}$ metric. One important reason is the failure when multiple objects correspond to one phrase. 
To decouple the effect of phrase extraction and REC, we design another baseline with spaCy and an ideal REC model named GoldREC. GoldREC outputs the ground-truth object whose corresponding phrase is the most similar to input phrases for any given input phrase. 
It shows that inaccurate phrase extraction has a large impact on final performance.

We also use MDETR~\cite{kamath2021mdetr} as a baseline. Our model outperforms MDETR on Flickr30k Entities with only half of the number of training epochs. It outperforms MDETR by $+5.8$ CMAP$_{50}$, demonstrating the effectiveness of decoupling image and text queries. 
We provide a visualization of these models' results in Appendix Table \ref{fig:flickr_vis}

\begin{table}[ht]
\begin{center}
\renewcommand{\arraystretch}{1.1}
\resizebox{\columnwidth}{!}{%
\begin{tabular}{lllll}
\shline
\textbf{Model} & \textbf{Pre-train Data} & \textbf{Epoches} & CMAP$_{50}$ & R@1  \\
\shline
OFA-REC+spaCy & CC, SBU, COCO, VG, OI, O365, YFCC (50M) & - & 23.2 & 58.1\\
GoldREC+spaCy & - & - & \color{gray}{44.4} & \color{gray}{100.0}\\
MDETR & COCO, VG, Flickr30k (200k) & 50 & 70.2 & 82.5 \\
\hline
DQ-DETR (Ours) & COCO, VG, Flickr30k (200k) & \textbf{25}  & \textbf{76.0} \small{\color{red}{(+5.8)}} & 83.2 \\
\shline
\end{tabular}}
\vspace{-0.2cm}
\caption{
\small{
Pre-training result comparison on Flickr30k Entities with three baselines: OFA-REC+spaCy~\cite{PengWang2022UNIFYINGAT, spacy2}, GoldREC+spaCy, and MDETR~\cite{kamath2021mdetr}. We assume GoldREC predicts accurate objects all the time. We mark the ideal results of GoldREC+spaCy in gray.
The numbers in brackets are the gains between DQ-DETR and MDETR.  All models are pre-trained with a ResNet-101 backbone. 
We use the notations ``CC'', ``SBU'', ``VG'', ``OI'', ``O365'', and ``YFCC'' for Conceptual Captions~\citep{PiyushSharma2018ConceptualCA}, SBU Captions~\citep{VicenteOrdonez2011Im2TextDI}, Visual Genome~\citep{RanjayKrishna2017VisualGC}, OpenImage~\cite{AlinaKuznetsova2018TheOI}, Objects365~\cite{zhou2019objects}, and YFCC100M~\cite{BartThomee2016YFCC100MTN} respectively. We use the \textsc{Any-Box} protocol for the ``R@1'' metric. 
}
}
\vspace{-0.2cm}
\label{tab:pretrain}
\end{center}
\end{table}

\subsubsection{4.2.2 Down-stream Task: Phrase Grounding}
\label{sec:phrase_grounding}
We compare our DQ-DETR with BAN~\cite{JinHwaKim2018BilinearAN}, VisualBert~\cite{LiunianHaroldLi2019VisualBERTAS}, CITE~\cite{BryanAPlummer2017ConditionalIE}, FAOG~\cite{ZhengyuanYang2019AFA}, SimNet-CCA~\cite{BryanAPlummer2020RevisitingIN}, DDPN~\cite{ZhouYu2018RethinkingDA}, RefTR~\cite{MuchenLi2021ReferringTA}, SeqTR~\cite{ChaoyangZhu2022SeqTRAS}, and MDETR~\cite{kamath2021mdetr} in Table \ref{tab:flickr}.
We fine-tune our pre-trained model on Flickr30k~\citep{BryanAPlummer2015Flickr30kEC} for the phrase grounding task. To compare with previous works in the literature, we follow MDETR~\citep{kamath2021mdetr} and evaluate the models with Recall@k under two different protocols, \textsc{Any-Box} and \textsc{Merged-Boxes} protocols.
For the \textsc{Any-Box} protocol, we evaluate our pre-trained model on the validation and test splits directly. For the \textsc{Merged-Boxes} protocol, we fine-tune the pre-trained model for $5$ epochs. Our model introduces improvements of $+0.7$ Recall@1 and $+1.4$ Recall@1 on the two validation splits, with only half of the number of pre-training epochs compared with MDETR. We also establish new state-of-the-art results on the two benchmarks with a ResNet-101 backbone.

\begin{table}[h]
\begin{center}
\small
\resizebox{0.85\columnwidth}{!}{%
\begin{tabular}{ccccccc}
 \shline
 Method & \multicolumn{3}{c}{Val} &  \multicolumn{3}{c}{Test} \\
  & R@1 & R@5 & R@10  & R@1 & R@5 & R@10 \\
 \shline
   \multicolumn{7}{c}{\textsc{Any-Box} Protocol}  \\
 \hline
 BAN  & - &  - & - & 69.7 &  84.2 &   86.4  \\
  VisualBert &  68.1 &  84.0 & 86.2 & - & - &  - \\
  VisualBert&  70.4 &  84.5 & 86.3 & 71.3 &  85.0 &  86.5  \\
  MDETR & {82.5} & {92.9} & {94.9}  & {83.4} & {93.5} & {95.3} \\
  DQ-DETR(Ours) & \textbf{83.2} \small{\color{red}{(+0.7)}} & \textbf{93.9} & \textbf{95.6}  & \textbf{83.9} & \textbf{94.6} & \textbf{96.2  } \\
   \shline
    \multicolumn{7}{c}{\textsc{Merged-Boxes} Protocol}  \\
   \hline
   CITE & - &  - & - & 61.9 &  - &   -  \\
   FAOG & - &  - & - & 68.7 &  - &   -  \\
   SimNet-CCA  & - &  - & - & 71.9 &  - &   -  \\
   DDPN  & 72.8 &  - & - & 73.5 &  - &   -  \\
   RefTR & - & - & - & 81.2 & - & - \\
   SeqTR & - & - & - & 81.2 & - & - \\
    MDETR & {82.3} & {91.8} & {93.7}  & {83.8} & {92.7} & {94.4} \\
    DQ-DETR(Ours) & \textbf{83.7} \small{\color{red}{(+1.4)}} & \textbf{93.8} & \textbf{95.8}  & \textbf{84.3} & \textbf{93.9} & \textbf{95.5} \\
 \shline
\end{tabular}}
\vspace{-0.28cm}
\caption{
\small{
Results on the phrase grounding task on Flickr30k Entities \cite{BryanAPlummer2015Flickr30kEC}.
All results are reported with a ResNet-101 backbone, except for SeqTR which uses DarkNet53~\cite{JosephRedmon2018YOLOv3AI}. We provide our gains compared with MDETR in brackets.}
\vspace{-0.3cm}
}
\label{tab:flickr}
\end{center}
\end{table}

\subsubsection{4.2.3 Down-stream Task: REC}
\label{sec:rec}

We compare our model with state-of-the-art REC methods on RefCOCO/+/g benchmarks after fine-tuning in Table. \ref{tab:refexp}. We evaluate the models with Recall@1. 
Although our model is not specifically designed for REC tasks, we can convert the REC task to a PEG problem by marking the whole sentence as a phrase corresponding to its referred object. 
As there are no ground truth phrases labeled in the dataset, we do not use the text mask-guided attention in the fine-tuning process.
As we have leveraged all training data of the three REC datasets during pre-training, it is reasonable to fine-tune the models on a combination of these three datasets. To avoid data leakage, we removed all images appeared in the val/test splits of RefCOCO/+/g. This operation removes about $10\%$ of the total images. 
As a result, our model outperforms all previous works with a ResNet-101 backbone and establishes new state-of-the-art results on the RefCOCO/+/g benchmarks. 

\begin{table*}[t]
\renewcommand{\arraystretch}{1.1}
\centering
\resizebox{0.96\textwidth}{!}{%
 \begin{tabular}{cccccccccc} 
 \shline
 Method  & Pre-training & \multicolumn{3}{c}{RefCOCO} &  \multicolumn{3}{c}{RefCOCO+} & \multicolumn{2}{c}{RefCOCOg}  \\ [0.5ex] 
         & image data  & val & testA & testB & val & testA & testB & val & test  \\
 \shline
 MAttNet~\citep{LichengYu2018MAttNetMA}  & None  &76.65  & 81.14 & 69.99 & 65.33 & 71.62 & 56.02 & 66.58 & 67.27  \\
 VGTR~\citep{YeDu2021VisualGW} & None  & 79.20 & 82.32 & 73.78 & 63.91 & 70.09 & 56.51 & 65.73 & 67.23 \\
 TransVG~\citep{JiajunDeng2021TransVGEV}  & None & 81.02 & 82.72 & 78.35 & 64.82 & 70.70 &  56.94 & 68.67 & 67.73 \\
 ViLBERT~\citep{JiasenLu2019ViLBERTPT}   & CC (3.3M) & - & -  & - & 72.34 & 78.52 &  62.61 & - & -   \\ 
 VL-BERT\_L~\citep{WeijieSu2019VLBERTPO}  & CC (3.3M) &- & -  & - & 72.59 & 78.57 &  62.30 & - & -\\
 UNITER\_L$^*$~\citep{YenChunChen2020UNITERLU}  & CC, SBU, COCO, VG (4.6M) & 81.41 & 87.04  & 74.17 & 75.90  & 81.45 & 66.70 & 74.86 & 75.77 \\  
 VILLA\_L$^*$~\citep{ZheGan2020LargeScaleAT}  & CC, SBU, COCO, VG (4.6M) & 82.39 & 87.48  & 74.84 &  76.17 & 81.54 &  66.84 & 76.18 & 76.71   \\ 
 ERNIE-ViL\_L~\citep{FeiYu2020ERNIEViLKE}  & CC, SBU (4.3M) &- & -  & - & 75.95 & 82.07 &  66.88 & - & -  \\ 
 RefTR~\citep{MuchenLi2021ReferringTA}  & VG (100k) & 85.65 & 88.73 & 81.16 & 77.55 & 82.26 & 68.99 & 79.25 & 80.01 \\
 SeqTR\dag~\cite{ChaoyangZhu2022SeqTRAS} & VG, COCO, Flickr30k, RIG (174k) &  87.00  & 90.15 &  \textbf{83.59} &  78.69 &  84.51  & 71.87  & 82.69  & 83.37 \\
 OFA~\cite{PengWang2022UNIFYINGAT} & CC, SBU, COCO, VG, OI, O365, YFCC (50M) & 88.48 & 90.67 & 83.30 & 81.39 & 87.15 & 74.29 & 82.29 & 82.31 \\
 MDETR~\citep{kamath2021mdetr}  & COCO, VG, Flickr30k (200k)  & {86.75} & {89.58}& {81.41}    & {79.52}  & {84.09} &  {70.62}  & {81.64} & {80.89}  \\
\hline
DQ-DETR (Ours) & COCO, VG, Flickr30k (200k) & \textbf{88.63}\small\color{red}{(+1.88)} & \textbf{91.04}\small\color{red}{(+1.46)} & {83.51}\small\color{red}{(+2.10)}    & \textbf{81.66}  & \textbf{86.15} &  \textbf{73.21}  & \textbf{82.76} & \textbf{83.44}  \\
\shline
\end{tabular}}
\vspace{-0.3cm}
\caption{
\small{
Top-1 accuracy comparison on the referring expression comprehension task. All models are trained with a ResNet-101 backbone, except for SeqTR~\cite{ChaoyangZhu2022SeqTRAS} which uses DarkNet53~\cite{JosephRedmon2018YOLOv3AI}. The models with $*$ may have a test data leakage issue for using a pre-trained BUTD detector~\citep{PeterAnderson2017BottomUpAT}. We use the notations ``CC'', ``SBU'', ``VG'', ``OI'', ``O365'', ``YFCC'', and ``RIG'' for Conceptual Captions~\citep{PiyushSharma2018ConceptualCA}, SBU Captions~\citep{VicenteOrdonez2011Im2TextDI}, Visual Genome~\citep{RanjayKrishna2017VisualGC}, OpenImage~\cite{AlinaKuznetsova2018TheOI}, Objects365~\cite{zhou2019objects}, YFCC100M~\cite{BartThomee2016YFCC100MTN}, and ReferItGame~\cite{SaharKazemzadeh2014ReferItGameRT} respectively.  
\dag Although using less images, SeqTR leveraged 6.1M image-text pairs for pre-training, which is more than our 1.3M training pairs.
We provide our gains compared with MDETR in brackets. Note that our models use only half of the number of pre-training epochs compared with MDETR.} 
}
\vspace{-0.25cm}
\label{tab:refexp}
\end{table*}

\subsubsection{4.2.4 Down-stream Task: DET}
To transfer our model to the standard object detection task, we reformulate the object detection task to a PEG task by concatenating the category names into a query text as in GLIP~\citep{li2021grounded}.
We fine-tune DQ-DETR and MDETR on the COCO detection-derived grounding data, and the results are shown in Table \ref{tab:coco_det}. We report the standard class-specific average precision for comparisons.
Our DQ-DETR outperforms MDETR with a large margin under the same setting. It also achieves comparable performance with main-stream detectors like Faster RCNN~\citep{ren2015faster} by training for longer epochs. Note that our model is not specifically designed for object detection tasks. Hence we do not use several techniques to boost performance like $300$ queries and focal loss~\citep{lin2018focal}, which may limit our performance on the task. 

\begin{table}[th]
    \vspace{-0.1cm}
    \centering
        \footnotesize
            \renewcommand{\arraystretch}{1.1}
    \resizebox{0.9\columnwidth}{!}{%
    \begin{tabular}{lclcccccc}
        \shline
        Model  & \#epochs & AP & AP$_{50}$ & AP$_{75}$ & AP$_{S}$ & AP$_{M}$ & AP$_{L}$  \\
        \shline
        Faster RCNN-FPN & 36 & 42.0 & 62.1 & 45.5 & 26.6 & 45.4 & 53.4  \\
        Faster RCNN-FPN & 108 & 44.0 & 63.9 & 47.8 & 27.2 & 48.1 & 56.0  \\
        \hline
        DETR            & 50 & 36.9 & 57.8 & 38.6 & 15.5 & 40.6 & 55.6 &  \\
        DETR            & 500 & 43.5 & 63.8 & 46.4 & 21.9 & 48.0 & 61.8 &  \\
        Conditional DETR & 50 & 42.8 & 63.7 & 46.0 & 21.7 & 46.6 & 60.9 \\
        \hline
        MDETR & (50+)12 & 36.5 &  57.2 &  37.9 & 14.7 &  40.0 & 57.5  \\
        DQ-DETR (Ours) & (25+)12 & 41.2 \small{{(+4.7)}} &  60.8 &  43.5 & 19.5 &  45.4 & 62.8 \\
        DQ-DETR (Ours) & (25+)50 & 42.8 \small{{(+6.3)}} &  63.0 &  45.2 & 21.8 &  47.4 & 63.3 \\
        \shline
    \end{tabular}
    }
    \vspace{-0.27cm}
    \caption{
    \small{
    Results for DQ-DETR and other models on the COCO~\citep{lin2015microsoft} object detection benchmark. Note that DQ-DETR is not designed for the object detection task. The numbers in black brackets are the number of pre-training epochs for each model. We provide our gains compared with MDETR in brackets. All models use a ResNet-101 backbone.
    }
    \vspace{-0.1cm}
    }
    \label{tab:coco_det}
    \vspace{-.1cm}
\end{table}

\subsection{The Base Setting \& Ablations}
\label{sec:base_setting}
We use MDETR~\citep{kamath2021mdetr} as our baseline and our model is an improvement upon it. 
As there are only pre-trained models in the MDETR paper, which is not easy for the community to compare, we design the base setting on Flickr30 Entities without pre-training in this section. All models are trained on $4$ Nvidia A100 GPUs with a ResNet-50 backbone and each GPU contains $4$ images.

\textbf{Results on Flickr30k Entities and ablations.} 
All models for Flickr30k are trained for $24$ epochs with a learning rate drop at the 20th epoch.
We compare our DQ-DETR and MDETR in Table \ref{tab:ablation}. Our proposed DQ-DETR outperforms MDETR with a large margin, e.g., +$13.14\%$ CMAP$_{50}$ and +$4.70\%$ Recall@1 on Flickr30k entities. 

\begin{table}[th]
    \centering
        \footnotesize
            \renewcommand{\arraystretch}{1.1}
    \resizebox{0.75\columnwidth}{!}{%
    \begin{tabular}{lcccc}
        \shline
        \textbf{Model} & CMAP$_{50}$ &  \textbf{R@$\textbf{1}$} & \textbf{R@$\textbf{5}$} & \textbf{R@$\textbf{10}$} \\        
        \shline
        MDETR  & $61.49$ & $77.46$ & $88.28$ & $91.19$\\
        MDETR \textit{with 200 queries}  & $56.89$ & $76.29$ & $86.87$ & $89.85$\\
        \hline
        Our baseline for DQ-DETR & $66.68$ & $75.44$& $87.94$ & $90.99$\\
        \quad + text mask attention & $68.26$ &  $76.58$ & $88.46$ & $91.31$ \\
        \quad + dual queries & $69.86$ &  $78.87$ & $89.39$ & $91.93$ \\
        \quad + positional query sharing & $\textbf{70.63}$ & $\textbf{79.16}$ & $\textbf{89.84}$ & $\textbf{92.10}$ \\
        \shline
    \end{tabular}
    }
    \vspace{-0.25cm}
    \caption{
    \small{Ablation results for DQ-DETR and a comparison with MDETR. All models are trained with a ResNet-50 backbone for 24 epochs, with a learning rate drop at the 20-th epoch. We use the \textsc{Any-Box} protocol for Recall@k metrics in this table.
    }}
    \label{tab:ablation}
    \vspace{-.2cm}
\end{table}

We provide the ablations in Table \ref{tab:ablation} as well. The dual query design introduces a gain of $1.60\%$ CMAP$_{50}$ in our experiments, which demonstrates the effectiveness of our dual query design. Moreover, we find the positional query sharing strategy helps improve the results as well, which highlights the necessity of our carefully designed dual queries. The multi-scale design and the text mask-guided attention also help the training of our model, each of which introduces a gain of $1-2\%$ CMAP$_{50}$.
We train a variant of MDETR with 200 queries for a fair comparison with our models. Surprisingly, we find the result drops a lot. We speculate that the data imbalance leads to the result, as more queries will exacerbate the imbalance of classes. In contrast, our model, which outputs 100 results like the original DETR, will not suffer from this problem. We will leave it as a future work to study how to scale up the model with more queries.

\section{Related Work}
\small{
\textbf{Referring expression comprehension (REC)} aims to predict the bounding boxes of the objects described by query texts. Classical methods can be divided into top-down methods (or two-stage methods) and bottom-up methods (or one-stage methods). The top-down methods~\citep{RichangHong2020LearningTC, RonghangHu2016ModelingRI, JingyuLiu2017ReferringEG, DaqingLiu2018LearningTA, VarunKNagaraja2016ModelingCB, JunhuaMao2015GenerationAC} view the task as a ranking task with a set of boxes predicted by a pre-trained object detector, which limits their performance. 
Most bottom-up solutions~\citep{XinpengChen2018RealTimeRE, YueLiao2019ARC,Miao2022ReferringEC, ZhengyuanYang2019AFA, ZhengyuanYang2020ImprovingOV} fuse multi-modality features and output the final boxes directly. Recently, DETR-like models~\citep{JiajunDeng2021TransVGEV, YeDu2021VisualGW, Miao2022ReferringEC, LiYang2022ImprovingVG} have attracted much attention for their simple architecture and promising results. \textbf{Phrase grounding} aims at associating regions to phrases. Early methods~\cite{LiweiWang2015LearningDS, LiweiWang2017LearningTN, BryanAPlummer2017ConditionalIE, HassanAkbari2018MultilevelMC, DaqingLiu2018LearningTA} learn to align cross-modal embeddings. The following methods leverage advanced models like RNN~\cite{RonghangHu2015NaturalLO, PelinDogan2019NeuralSP}, Transformer~\cite{LiunianHaroldLi2020WhatDB, JiaboYe2022ShiftingMA, JiahaoLi2022AdaptingCF}, and graph architectures~\cite{MohitBajaj2019G3raphGroundGL, YongfeiLiu2019LearningCC, ZongshenMu2021DisentangledMG} to capture context information. Despite their encouraging progress, most models rely on phrase annotations during inference. \\
\textbf{Object detection} is a fundamental task in vision and has a wide influence on phrase grounding. DETR~\citep{carion2020end} introduces an end-to-end Transformer~\citep{vaswani2017attention}-based solution for the first time. 
Many follow-ups~\citep{zhu2020deformable, meng2021conditional, gao2021fast, Dai_2021_ICCV, wang2021anchor, liu2022dabdetr, li2022dn, zhang2022dino} improve the model performance and training efficiency of DETR. However, the methods have no perceptions of languages. \\
\textbf{Visual-language pre-trained models} have dominated the vision-language field recently. However, most fine-grained visual-language models~\citep{YenChunChen2020UNITERLU, PengchuanZhang2021VinVLRV, li2020oscar, LiunianHaroldLi2019VisualBERTAS, HaoTan2019LXMERTLC, JiasenLu2019ViLBERTPT, JiasenLu201912in1MV, WeijieSu2019VLBERTPO} rely on pre-trained object detectors for region feature extraction. MDETR~\citep{kamath2021mdetr} proposed to train a visual-language models with detection simultaneously. After MDETR, GLIP~\citep{li2021grounded} proposes a contrastive learning to unify object detection and visual grounding. Despite their promising performance, they normally treat phrase grounding as an extended task of object detection and overlook the different needs of two different tasks: box regression and phrase localization. In contrast, we follow MDETR and design an end-to-end model for PEG pre-training, yielding a more effective and principled solution than previous works. 
}

\section{Conclusion}
We have presented an overview of visual grounding tasks and identified an often overlooked phrase extraction step. The observation inspires us to re-emphasize a PEG (phrase extraction and grounding) task and propose a new CMAP (cross-modal average precision) metric. The CMAP overcomes the ambiguity of Recall@1 in many-box-to-one-phrase cases in phrase grounding tasks. 
Moreover, we propose a new interpretation of the PEG task as a dual detection problem by viewing phrase localization as a 1D text segmentation problem. With this new interpretation, we developed a dual query-based DETR-like model DQ-DETR for phrase grounding. Such a decoupled query design helps alleviate the difficulty of modality alignment between image and text, yielding both faster convergence and better performance. We also proposed a text mask-guided attention to constrain a text query to the masked text tokens in cross-attention modules. We conducted extensive experiments to verify the effectiveness of our model design. \\

\section{Acknowledgement}
This work was supported by the National Key Research and Development Program of China (2020AA0106302). \\
We thank Yukai Shi, Linghao Chen, Jianan Wang, Ailing Zeng, and Xianbiao Qi of IDEA CVR groups for their valuable feedbacks. We thank all the reviewers including SPC and AC in AAAI 2023 for their kindly suggestions. We thank the reviewers of our initial version in NeurIPS 2022 for their valuable suggestions, especially the suggestions of the Reviewer btPY, which helped us a lot.

\bibliography{aaai23}


\clearpage

\appendix

\section{Dual Query Design}
\label{sec:appendix_dual_query}
\begin{figure}[h]
\centering
\resizebox{\columnwidth}{!}{%
\includegraphics{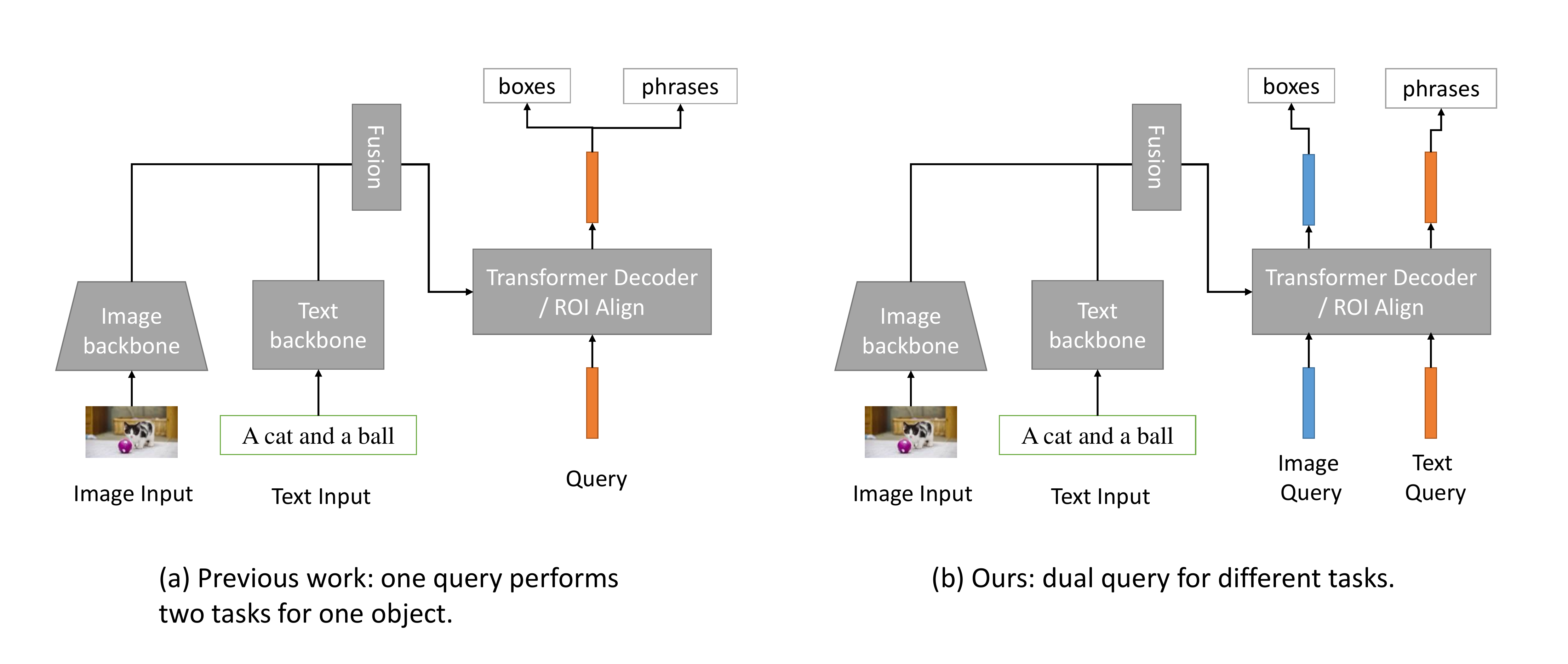}}
\vspace{-0.6cm}
\caption{Comparison of our dual query design and previous designs.}
\label{fig:appendix_dual_query}
\end{figure}

We provide a comparison between our dual query design and previous single query design in Fig. \ref{fig:appendix_dual_query}. We abstract all previous work in Fig. \ref{fig:appendix_dual_query} (a), where each query corresponds to one object. Based on the novel view for the phrase grounding task by formulating it as a dual detection problem, we propose to use dual queries for text and image information separately. 

\section{More Implementation Details}
\label{sec:imple_details}

\textit{Training.} Following MDETR~\citep{kamath2021mdetr}, we augment the training images with random resize and random crop. The max length of the longer side of images is $1333$. We use AdamW~\citep{DiederikPKingma2014AdamAM, loshchilov2019decoupled} with weight decay $1e^{-4}$ during training. 
\noindent
\textit{Datasets:} \textit{Flickr30k Entities}~\citep{BryanAPlummer2015Flickr30kEC} contains 31,783 images in total, including 159k caption sentences and 427k annotated phrases. We use splits from MDETR, which contains 149k training samples and about 5k validation samples. \\
\textit{RefCOCO/+/g} refers to three datasets, RefCOCO, RefCOCO+, and RefCOCOg, for the REC task. They use the images and referred objects from COCO~\citep{lin2015microsoft}. We report scores on the validation, testA, and testB splits on RefCOCO and RefCOCO+. For RefCOCOg, we follow \citep{VarunKNagaraja2016ModelingCB} to split the dataset into train, validation, and test. For the combination of RefCOCO/+/g, we concatenate the train splits of RefCOCO/+/g after removing the images that appear in the validation or test splits of any dataset. 
\\
\textit{COCO 2017 Detection.} We use the standard COCO 2017 dataset for the object detection task. The COCO 2017 detection dataset contains 118k training images and 5k validation images.

\section{Hyper-Parameters}
We list the main hyper-parameters in Table. \ref{sec:hyper_parameters} to help others reproduce our results. We will release the code after the blind review.
\label{sec:hyper_parameters}

\begin{table}[ht]
    \centering
    \begin{tabular}{l|c}
	\shline
	Item & Value \\
	\shline
	lr & 0.0001 \\ \hline
	lr\_backbone & 1e-05 \\ \hline
	text\_encoder\_lr & 1e-05 \\ \hline
	fraction\_warmup\_steps & 0.01 \\ \hline
	weight\_decay & 0.0001 \\ \hline
	clip\_max\_norm & 0.1 \\ \hline
	enc\_layers & 6 \\ \hline
	dec\_layers & 6 \\ \hline
	dim\_feedforward & 2048 \\ \hline
	hidden\_dim & 256 \\ \hline
	dropout & 0.1 \\ \hline
	nheads & 8 \\ \hline
	num\_queries & 100 \\ \hline
	transformer\_activation & ``relu'' \\ \hline
	batch\_norm\_type & ``FrozenBatchNorm2d'' \\ \hline
	set\_cost\_class & 1.0 \\ \hline
	set\_cost\_bbox & 5.0 \\ \hline
	set\_cost\_giou & 2.0 \\ \hline
	ce\_loss\_coef & 2.0 \\ \hline
	bbox\_loss\_coef & 5.0 \\ \hline
	giou\_loss\_coef & 2.0 \\ \hline
	cls\_temperature & 0.07\\ \hline
	pre\_norm &  \texttt{False} \\ \hline
	\shline
    \end{tabular}
    \vspace{0.2cm}
    \caption{Hyper-parameters used in our pre-trained models.}
    \label{tab:hyperparameters}
\end{table}

\section{Inference Speed}
We present the inference time of our DQ-DETR and MDETR in Table \ref{tab:fps}. Our DQ-DETR has a little more computational cost compared with MDETR, but not a significantly burden.

\begin{table}[ht]
\centering
\renewcommand{\arraystretch}{1.1}
\resizebox{0.8\columnwidth}{!}{%
 \begin{tabular}{lc} 
 \shline
 Method   & inference speed(fps) \\
 \shline
 MDETR   & 26.7 \\
 DQ-DETR (Ours)   & 20.4 \\
 \shline
\end{tabular}}
\vspace{0.1cm}
\caption{Comparisons of the inference time of our DQ-DETR and MDETR.}
\label{tab:fps}
\end{table}

\section{Visualizations}
\subsection{Comparisons on Flickr30k Entities for PEG}
\label{sec:weaknesses_two_stage}
\begin{figure*}[htbp]
\centering
\resizebox{1.0\textwidth}{!}{%
\subfigure{
\begin{minipage}[t]{1.0\linewidth}
    \subfigure{
    \begin{minipage}[t]{0.22\linewidth}
    \centering
    GT
    \end{minipage}
    }
    \subfigure{
    \begin{minipage}[t]{0.22\linewidth}
    \centering
    OFA-REC+spaCy
    \end{minipage}
    }
    \subfigure{
    \begin{minipage}[t]{0.22\linewidth}
    \centering
    MDETR
    \end{minipage}
    }
    \subfigure{
    \begin{minipage}[t]{0.22\linewidth}
    \centering
    DQ-DETR(Ours)
    \end{minipage}
    }
\vspace{-1.0cm}
\end{minipage}
}}
\resizebox{1.0\textwidth}{!}{%
\subfigure{
\begin{minipage}[t]{1.0\linewidth}
    \subfigure{
    \begin{minipage}[t]{0.22\linewidth}
    \centering
    \includegraphics[width=\linewidth]{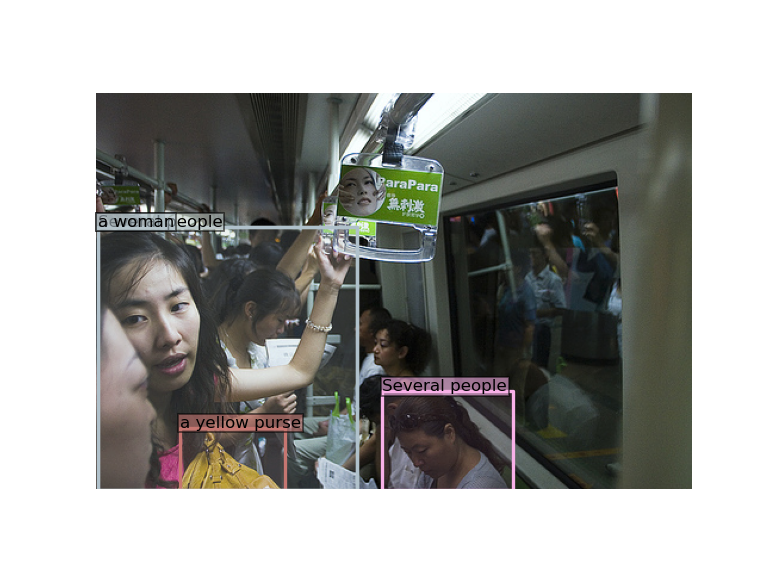}
    \end{minipage}
    }
    \subfigure{
    \begin{minipage}[t]{0.22\linewidth}
    \centering
    \includegraphics[width=\linewidth]{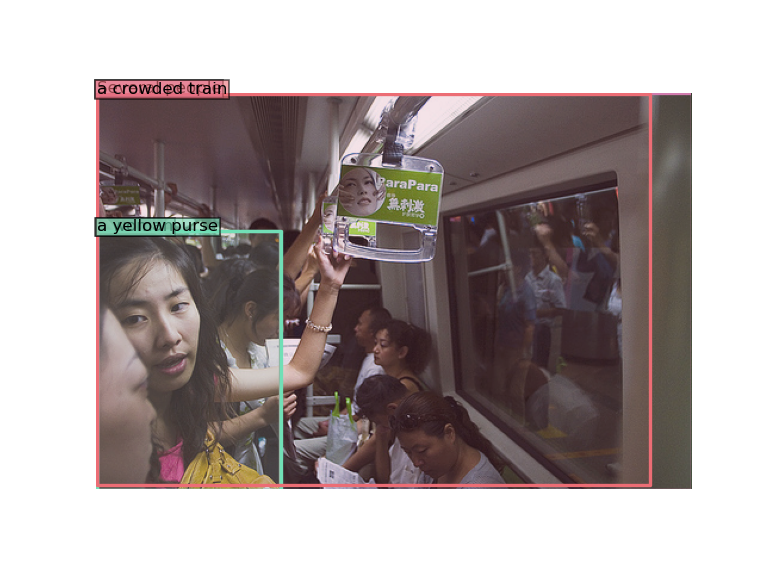}
    \end{minipage}
    }
    \subfigure{
    \begin{minipage}[t]{0.22\linewidth}
    \centering
    \includegraphics[width=\linewidth]{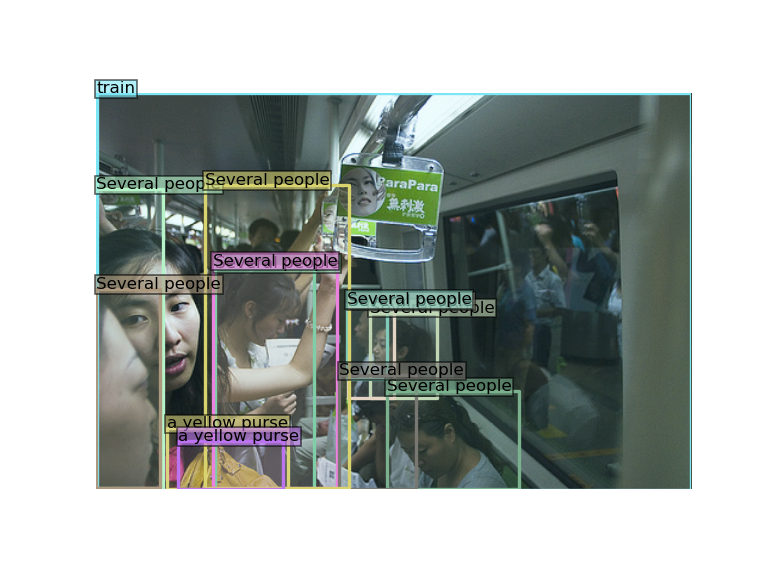}
    \end{minipage}
    }
    \subfigure{
    \begin{minipage}[t]{0.22\linewidth}
    \centering
    \includegraphics[width=\linewidth]{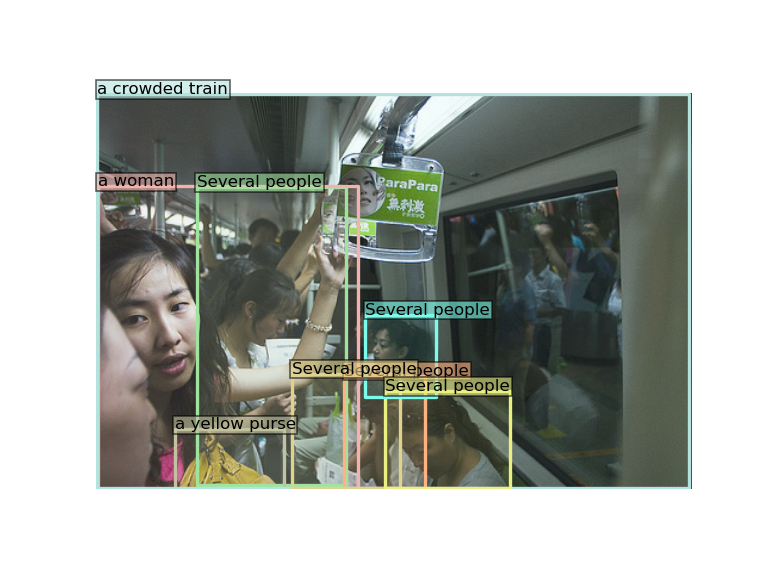}
    \end{minipage}
    }
\end{minipage}
}}
\resizebox{0.9\textwidth}{!}{\subfigure{
\begin{minipage}[t]{0.9\linewidth}
    \vspace{-0.8cm}
    {Caption: \texttt{Several people including a woman with a yellow purse are riding a crowded train .}}
\end{minipage}
}}
\resizebox{1.0\textwidth}{!}{%
\subfigure{
\begin{minipage}[t]{1.0\linewidth}
    \subfigure{
    \begin{minipage}[t]{0.22\linewidth}
    \centering
    \includegraphics[width=\linewidth]{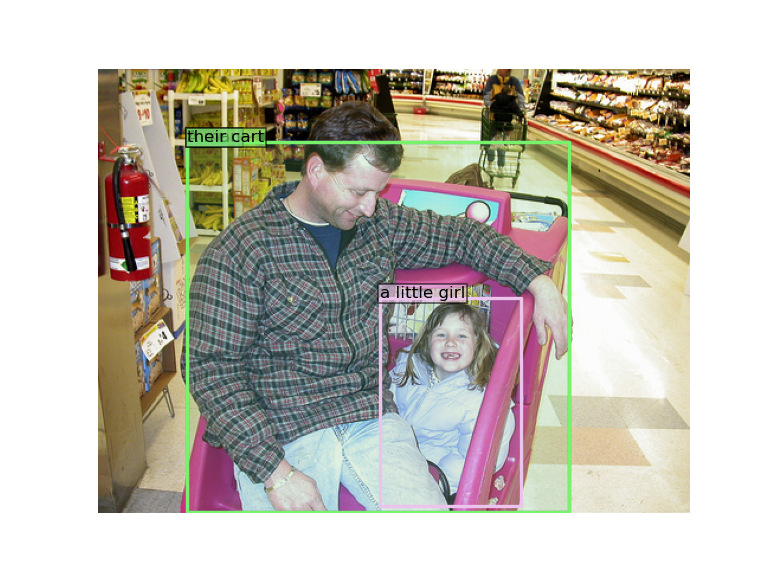}
    \end{minipage}
    }
    \subfigure{
    \begin{minipage}[t]{0.22\linewidth}
    \centering
    \includegraphics[width=\linewidth]{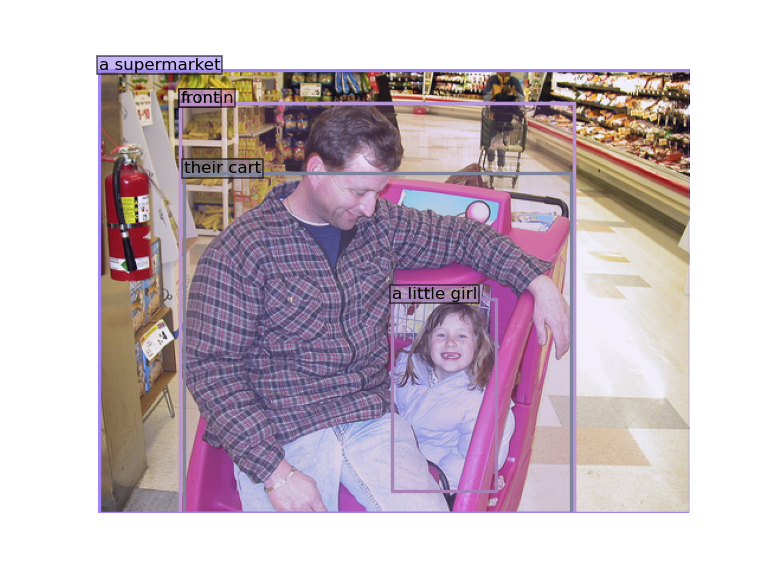}
    \end{minipage}
    }
    \subfigure{
    \begin{minipage}[t]{0.22\linewidth}
    \centering
    \includegraphics[width=\linewidth]{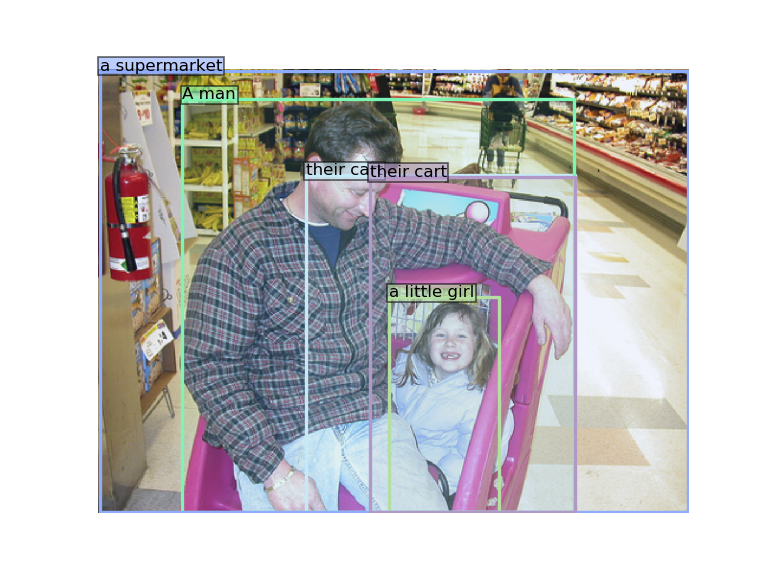}
    \end{minipage}
    }
    \subfigure{
    \begin{minipage}[t]{0.22\linewidth}
    \centering
    \includegraphics[width=\linewidth]{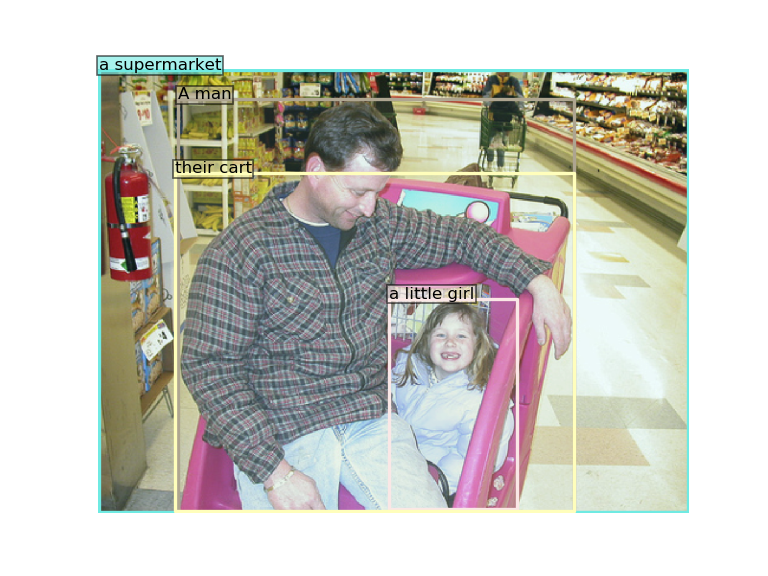}
    \end{minipage}
    }
\end{minipage}
}}
\resizebox{0.9\textwidth}{!}{\subfigure{
\begin{minipage}[t]{0.9\linewidth}
    \vspace{-0.8cm}
    {Caption: \texttt{A man and a little girl happily posing in front of their cart in a supermarket .}}
\end{minipage}
}}
\resizebox{1.0\textwidth}{!}{%
\subfigure{
\begin{minipage}[t]{1.0\linewidth}
    \subfigure{
    \begin{minipage}[t]{0.22\linewidth}
    \centering
    \includegraphics[width=\linewidth]{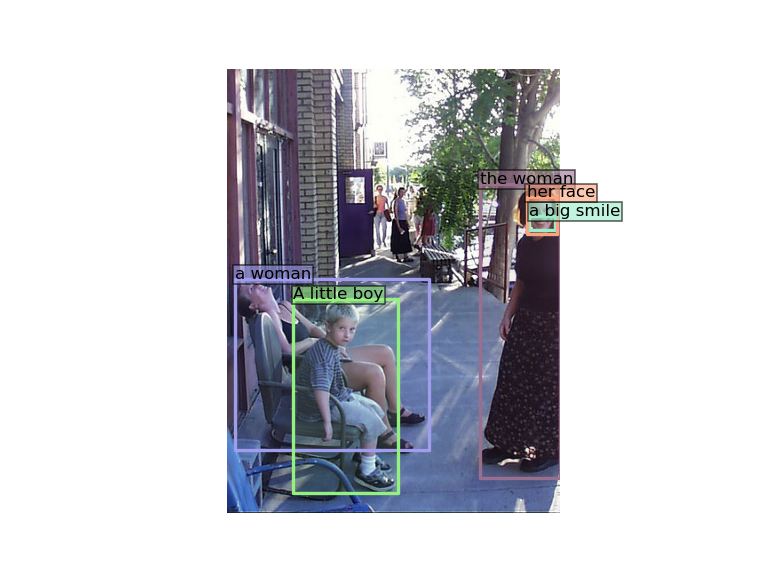}
    \end{minipage}
    }
    \subfigure{
    \begin{minipage}[t]{0.22\linewidth}
    \centering
    \includegraphics[width=\linewidth]{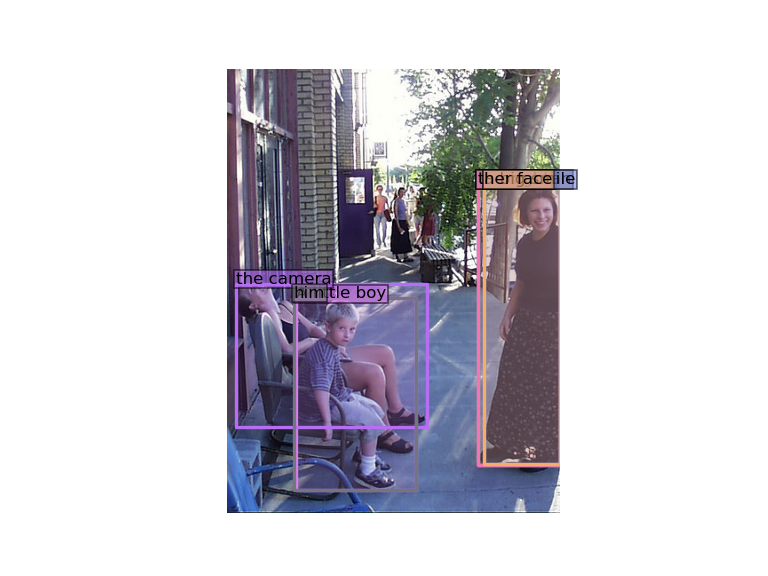}
    \end{minipage}
    }
    \subfigure{
    \begin{minipage}[t]{0.22\linewidth}
    \centering
    \includegraphics[width=\linewidth]{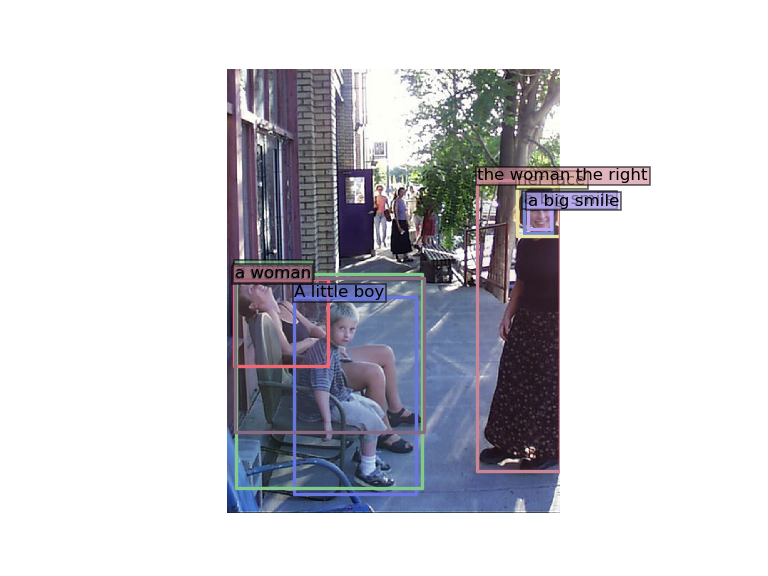}
    \end{minipage}
    }
    \subfigure{
    \begin{minipage}[t]{0.22\linewidth}
    \centering
    \includegraphics[width=\linewidth]{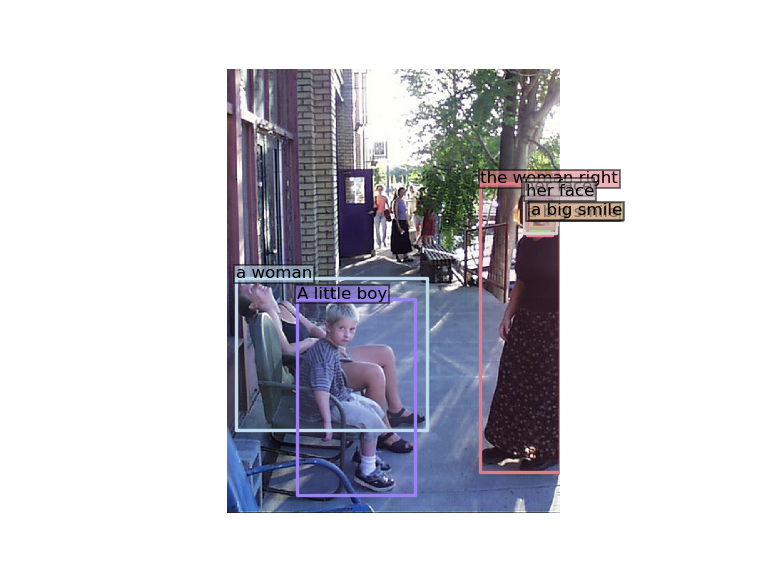}
    \end{minipage}
    }
\end{minipage}
}}
\resizebox{0.9\textwidth}{!}{\subfigure{
\begin{minipage}[t]{0.9\linewidth}
    \vspace{-0.8cm}
    {Caption: \texttt{A little boy looks at the camera while a woman behind him seems to be laughing very hard and the woman on the right has a big smile on her face .}}
\end{minipage}
}}
\resizebox{1.0\textwidth}{!}{%
\subfigure{
\begin{minipage}[t]{1.0\linewidth}
    \subfigure{
    \begin{minipage}[t]{0.22\linewidth}
    \centering
    \includegraphics[width=\linewidth]{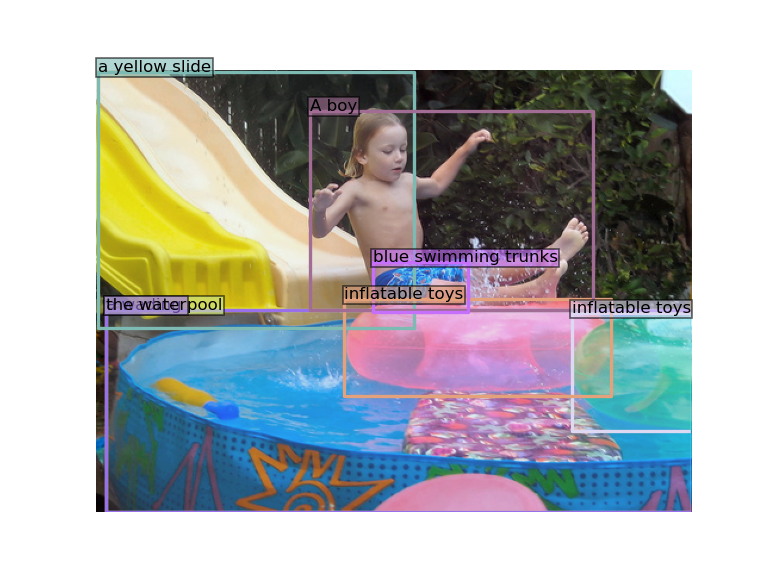}
    \end{minipage}
    }
    \subfigure{
    \begin{minipage}[t]{0.22\linewidth}
    \centering
    \includegraphics[width=\linewidth]{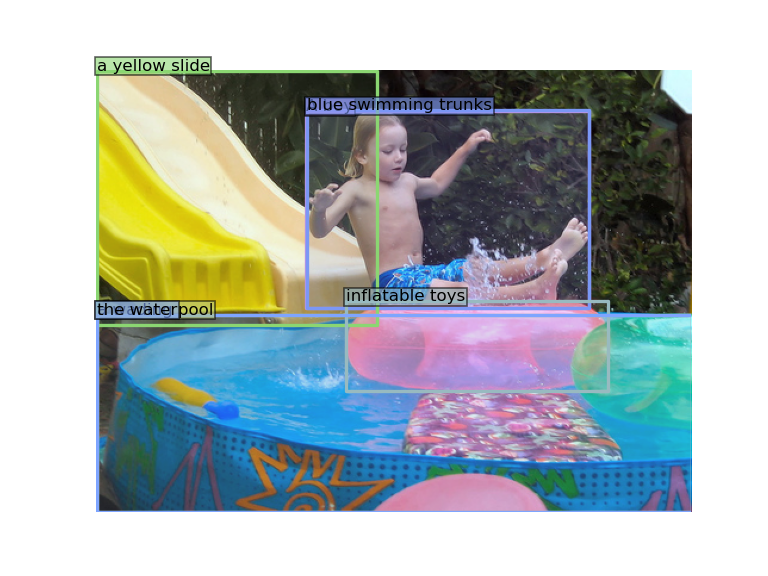}
    \end{minipage}
    }
    \subfigure{
    \begin{minipage}[t]{0.22\linewidth}
    \centering
    \includegraphics[width=\linewidth]{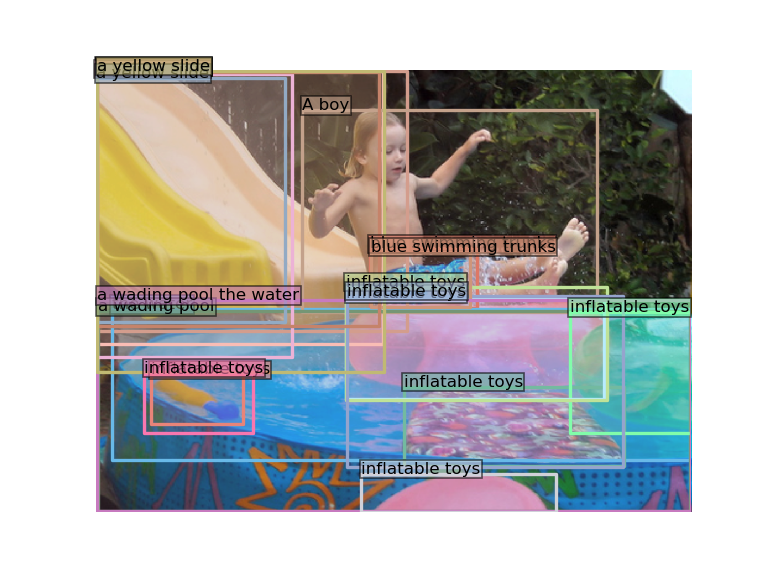}
    \end{minipage}
    }
    \subfigure{
    \begin{minipage}[t]{0.22\linewidth}
    \centering
    \includegraphics[width=\linewidth]{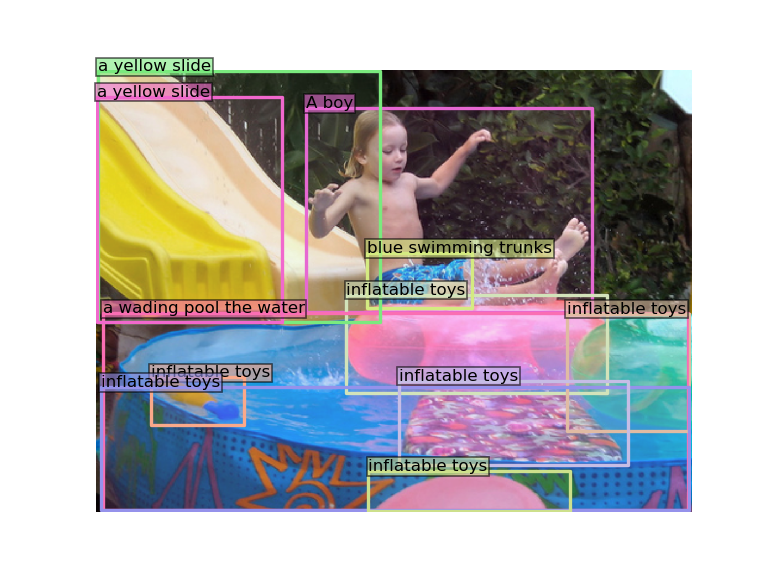}
    \end{minipage}
    }
\end{minipage}
}}
\resizebox{0.9\textwidth}{!}{\subfigure{
\begin{minipage}[t]{0.9\linewidth}
    \vspace{-0.8cm}
    {Caption: \texttt{A boy in blue swimming trunks slides down a yellow slide into a wading pool with inflatable toys floating in the water .}}
\end{minipage}
}}
\resizebox{1.0\textwidth}{!}{%
\subfigure{
\begin{minipage}[t]{1.0\linewidth}
    \subfigure{
    \begin{minipage}[t]{0.22\linewidth}
    \centering
    \includegraphics[width=\linewidth]{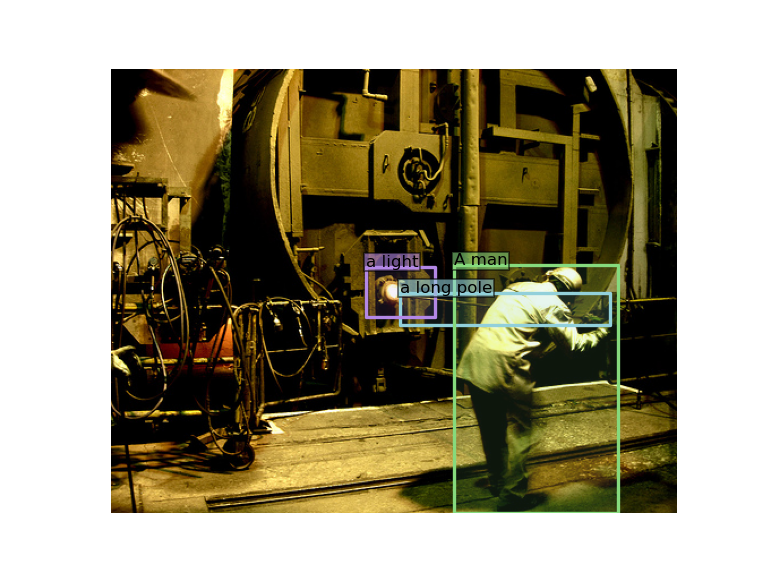}
    \end{minipage}
    }
    \subfigure{
    \begin{minipage}[t]{0.22\linewidth}
    \centering
    \includegraphics[width=\linewidth]{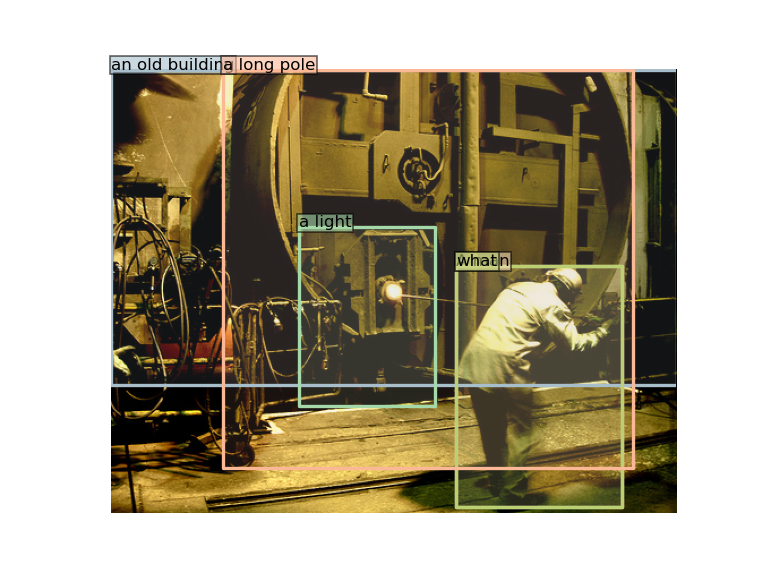}
    \end{minipage}
    }
    \subfigure{
    \begin{minipage}[t]{0.22\linewidth}
    \centering
    \includegraphics[width=\linewidth]{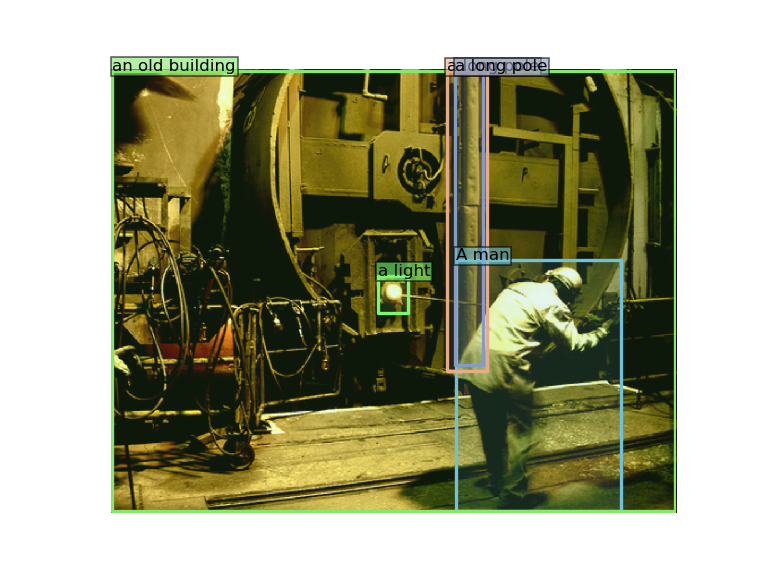}
    \end{minipage}
    }
    \subfigure{
    \begin{minipage}[t]{0.22\linewidth}
    \centering
    \includegraphics[width=\linewidth]{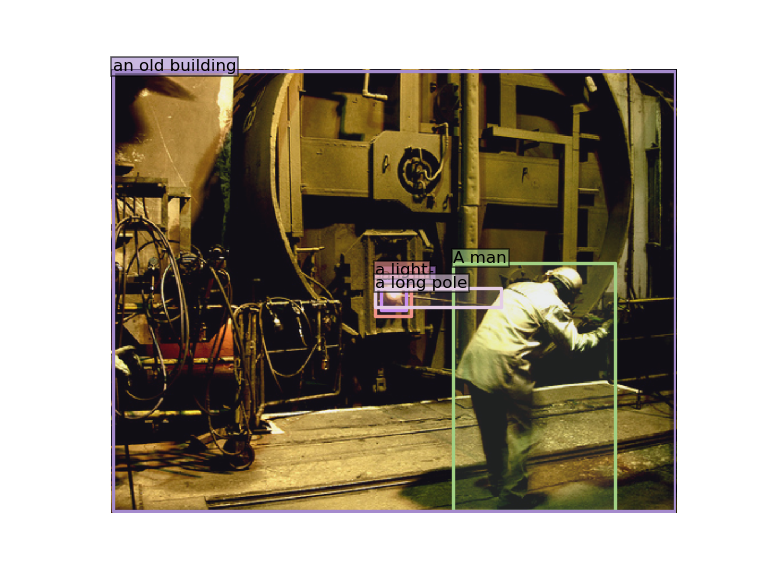}
    \end{minipage}
    }
\end{minipage}
}}
\resizebox{0.9\textwidth}{!}{\subfigure{
\begin{minipage}[t]{0.9\linewidth}
    \vspace{-0.8cm}
    {Caption: \texttt{A man in an old building holding what looks like a light on a long pole .}}
\end{minipage}
}}
\centering
\caption{Comparisons of PEG results on Flickr30k entities~\cite{BryanAPlummer2015Flickr30kEC}. We compare our model with OFA-REC~\cite{PengWang2022UNIFYINGAT}+sapCy~\cite{spacy2} and MDETR~\cite{kamath2021mdetr}. ``GT'' means the grounding truth results. }
\label{fig:flickr_vis}
\end{figure*}

We compare the results of our DQ-DETR with two competitors, MDETR~\cite{kamath2021mdetr} and OFA-REC+spaCy~\cite{PengWang2022UNIFYINGAT, spacy2} in Fig. \ref{fig:flickr_vis}. 

The visualizations of OFA-REC+spaCy present some examples of the weaknesses of two-stage models we stated in Sec \ref{sec:intro}:
\begin{enumerate}
    \item The REC model can only predict one object for each phrase, as shown in the first row of Fig. \ref{fig:flickr_vis}. For the phrase input \texttt{Several people}, the model outputs a combination of these people. While for the phrase \texttt{inflatable toys} in the fourth row, the model outputs only one of the ground truth boxes.
    \item The spaCy may extract phrase unrelated to the input images, like the phrase \texttt{camera} in the third row of Fig. \ref{fig:flickr_vis}. 
    \item The spaCy may extract inaccuracy phrases, which interferences the performance of REC models. For example, the extracted phrases \texttt{front} in the second row and \texttt{what} in the fifth row of the Fig. \ref{fig:flickr_vis} is meaningless. The phrase \texttt{a woman behind him} in the third row of Fig. \ref{fig:flickr_vis} are not exactly extracted by spaCy.
    \item The REC model predicts inaccuracy boxes, like boxes of the phrase \texttt{a yellow purse} in the first row and the phrase \texttt{her face} in the third row of Fig. \ref{fig:flickr_vis}.
\end{enumerate}

The comparisons between MDETR and our DQ-DETR show that our model has better abilities for both object localization and phrase extractions.

\subsection{Comparisons on RefCOCO for REC}
\begin{figure*}[htbp]
\centering
\resizebox{0.8\textwidth}{!}{%
\subfigure{
\begin{minipage}[t]{0.5\linewidth}
\centering
\includegraphics[width=\linewidth]{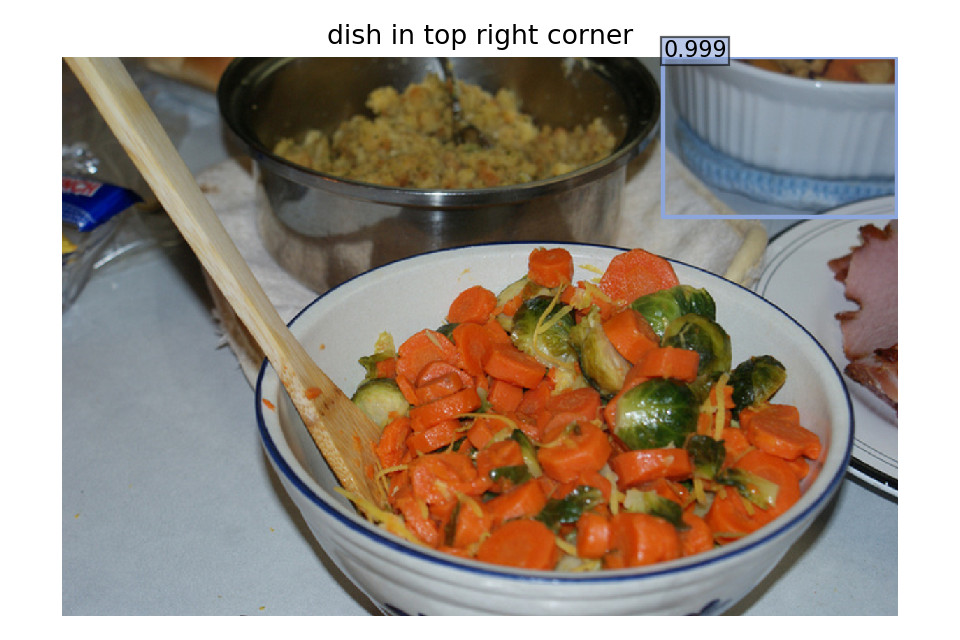}
\end{minipage}
}%
\subfigure{
\begin{minipage}[t]{0.5\linewidth}
\centering
\includegraphics[width=\linewidth]{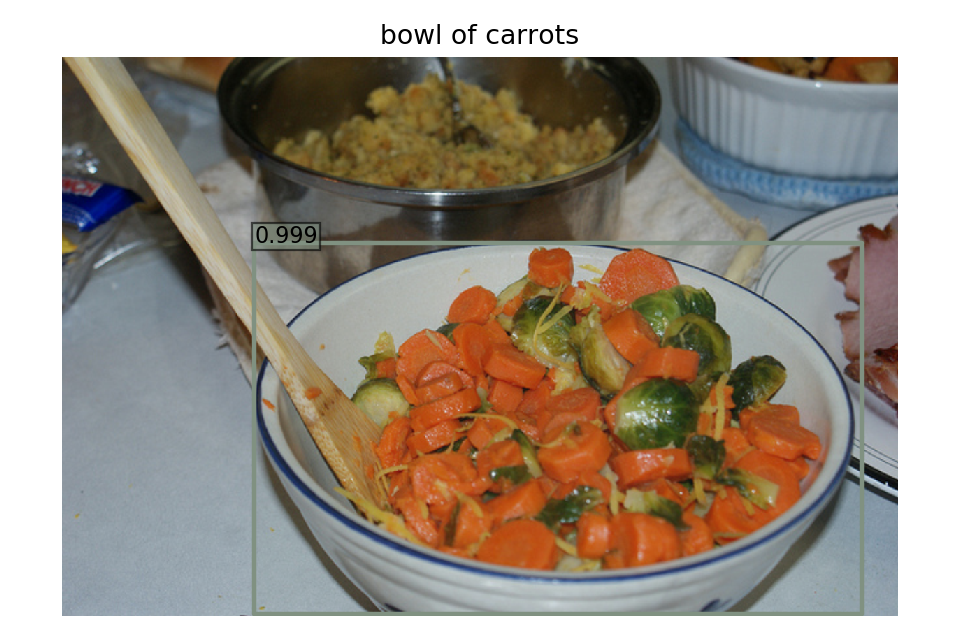}
\end{minipage}
}}
\resizebox{0.8\textwidth}{!}{%
\subfigure{
\begin{minipage}[t]{0.5\linewidth}
\centering
\includegraphics[width=\linewidth]{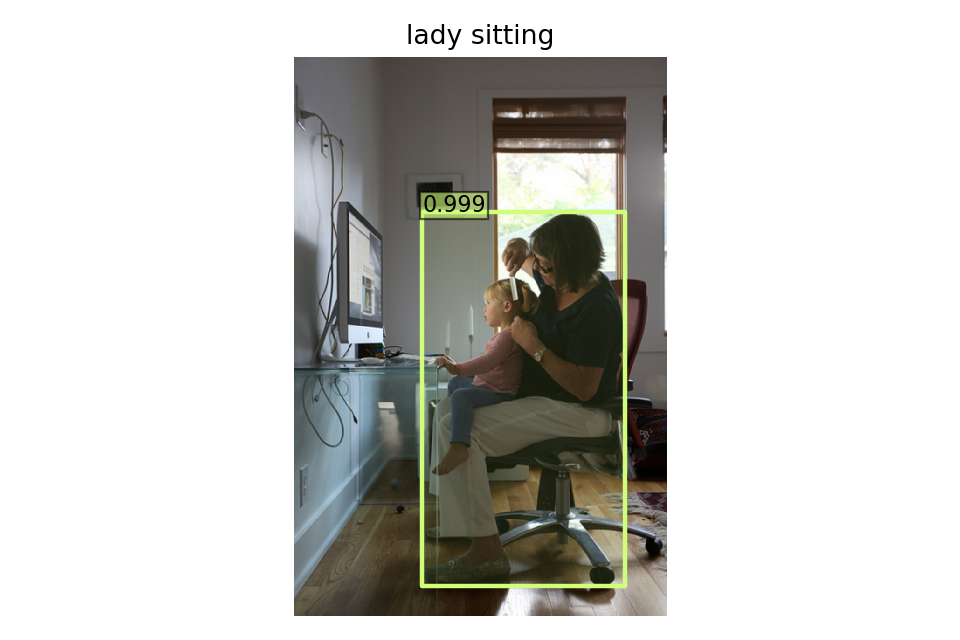}
\end{minipage}
}%
\subfigure{
\begin{minipage}[t]{0.5\linewidth}
\centering
\includegraphics[width=\linewidth]{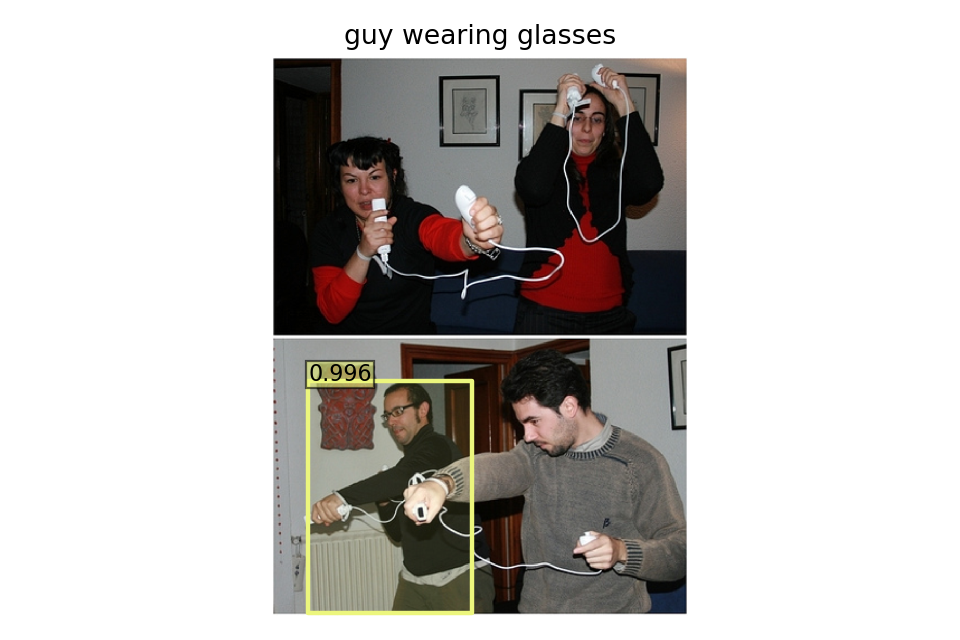}
\end{minipage}
}}
\resizebox{0.8\textwidth}{!}{%
\subfigure{
\begin{minipage}[t]{0.5\linewidth}
\centering
\includegraphics[width=\linewidth]{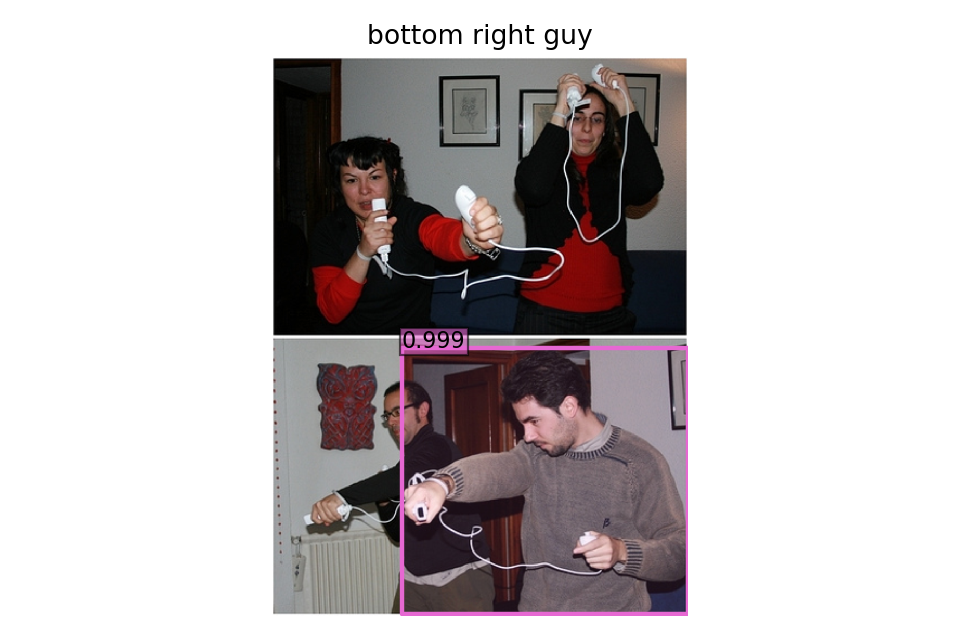}
\end{minipage}
}%
\subfigure{
\begin{minipage}[t]{0.5\linewidth}
\centering
\includegraphics[width=\linewidth]{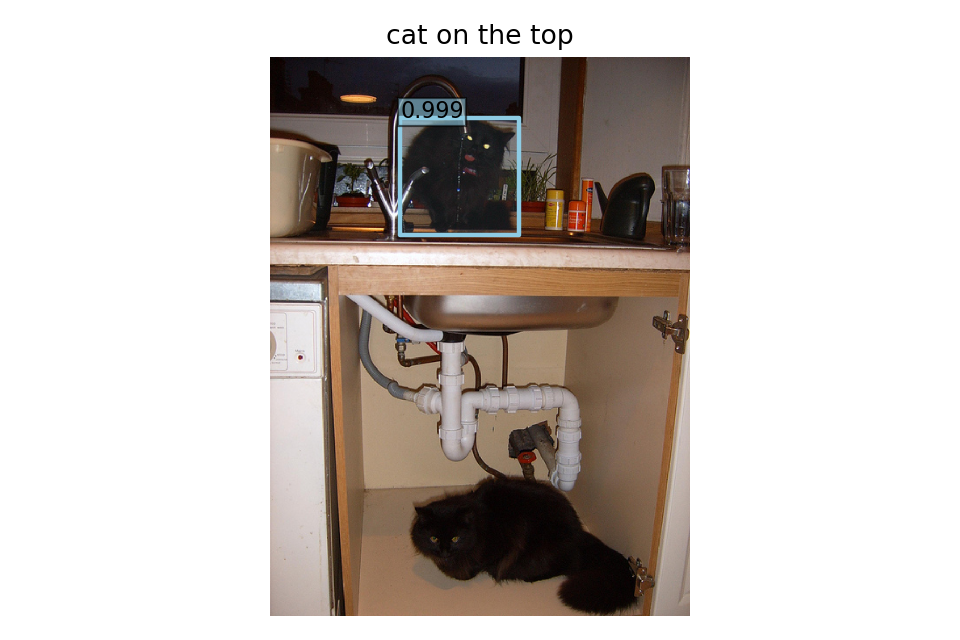}
\end{minipage}
}}
\resizebox{0.8\textwidth}{!}{%
\subfigure{
\begin{minipage}[t]{0.5\linewidth}
\centering
\includegraphics[width=\linewidth]{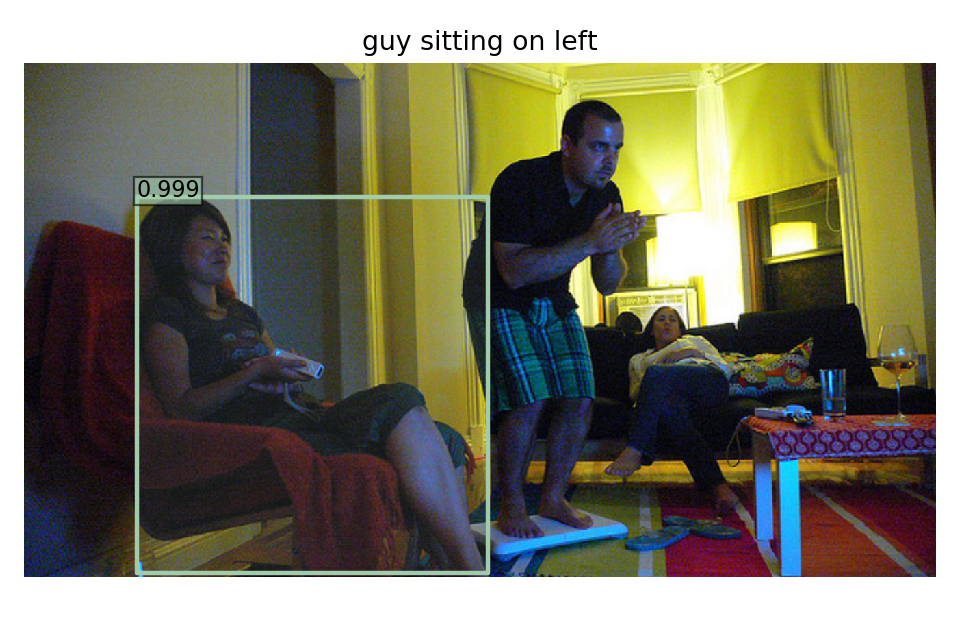}
\end{minipage}
}%
\subfigure{
\begin{minipage}[t]{0.5\linewidth}
\centering
\includegraphics[width=\linewidth]{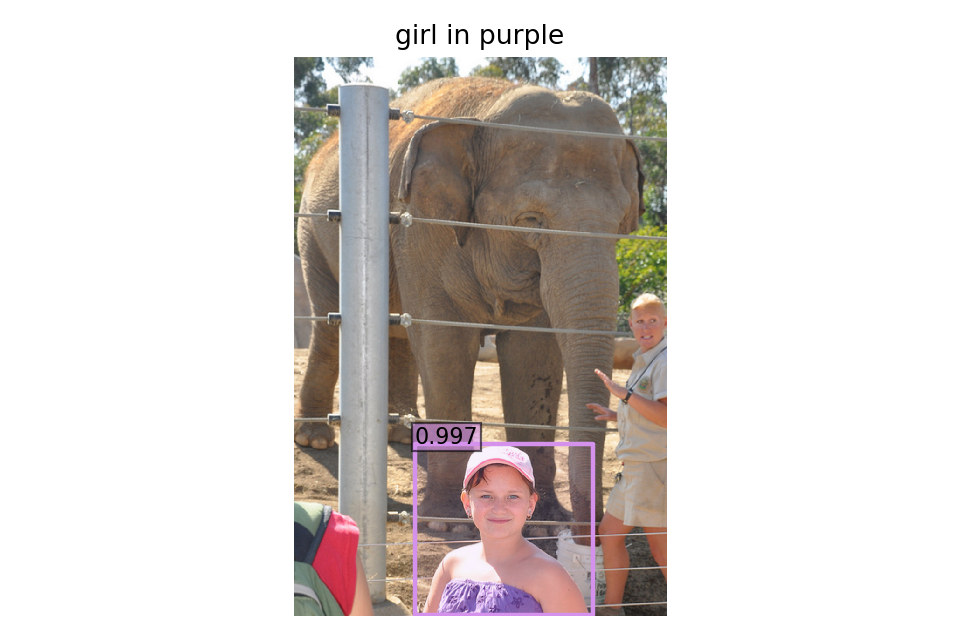}
\end{minipage}
}}
\centering
\caption{Visualizations of bounding box outputs  of our DQ-DETR on the RefCOCO dataset. Captions above images are input texts, and numbers on the bounding boxes are prediction scores.
}
\label{fig:refcoco_vis}
\end{figure*}

We provide some visualization of our model on RefCOCO in Fig. \ref{fig:refcoco_vis}.

\subsection{Comparisons on COCO Detection}
\begin{figure*}[htbp]
\centering
\resizebox{0.9\textwidth}{!}{%
\subfigure{
\begin{minipage}[t]{0.5\linewidth}
    \subfigure{
    \begin{minipage}[t]{1.0\linewidth}
    \centering
    \includegraphics[width=\linewidth]{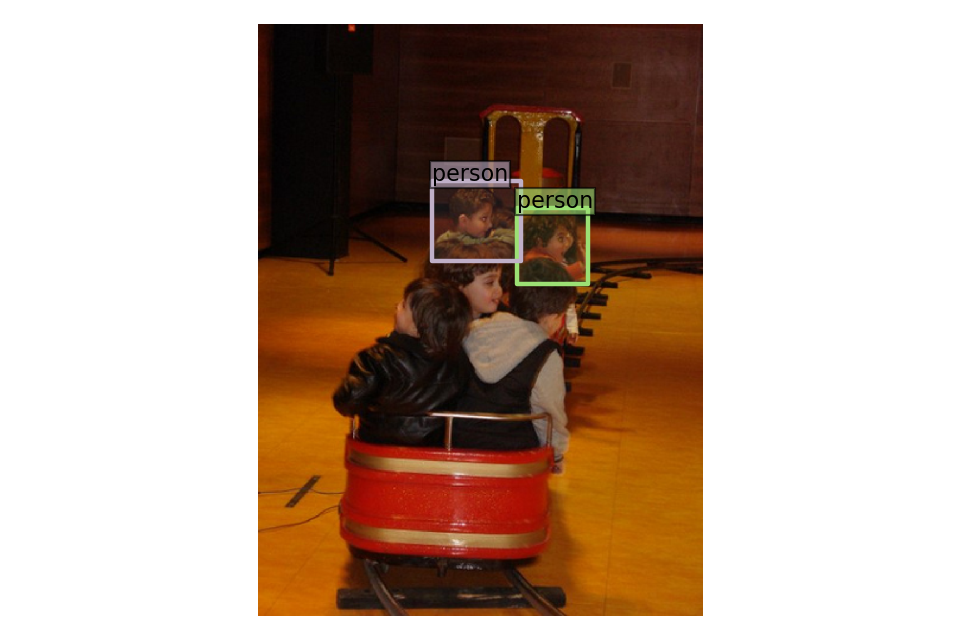}
    \end{minipage}
    }
    \subfigure{
    \begin{minipage}[t]{1.0\linewidth}
    \centering
    \includegraphics[width=\linewidth]{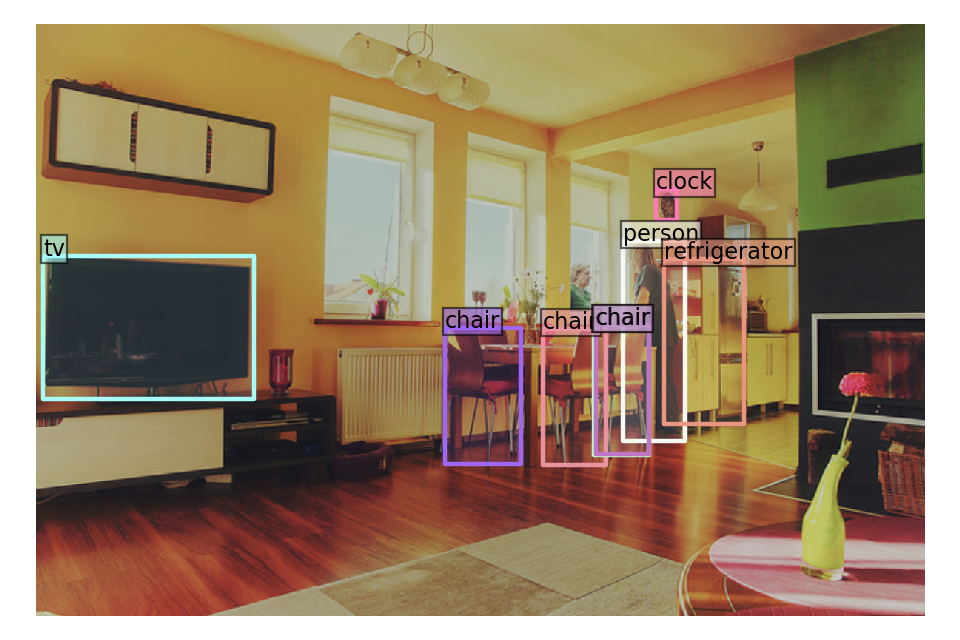}
    \end{minipage}
    }
    \subfigure{
    \begin{minipage}[t]{1.0\linewidth}
    \centering
    \includegraphics[width=\linewidth]{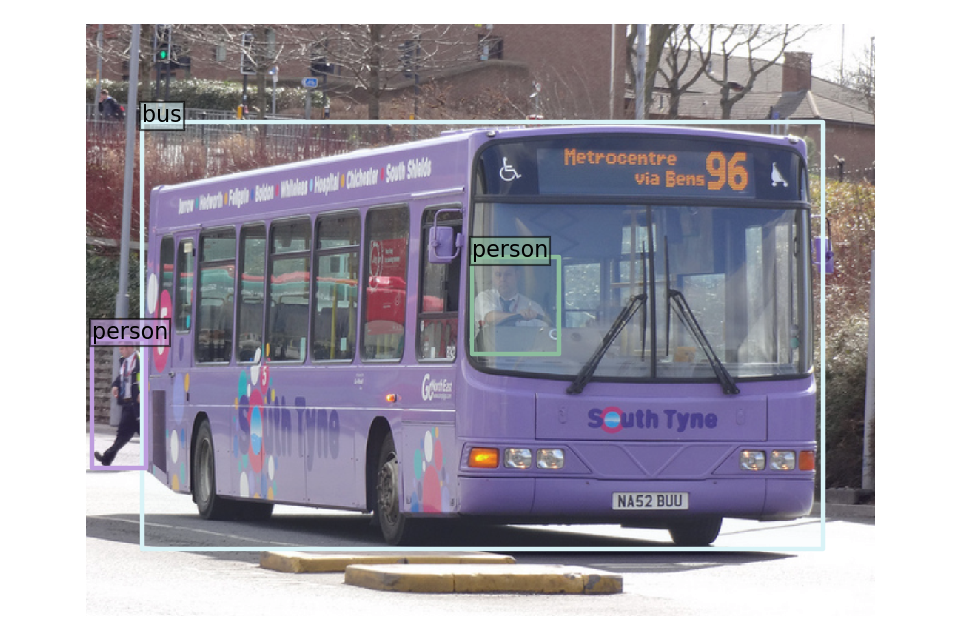}
    \end{minipage}
    }
\end{minipage}
}%
\subfigure{
\begin{minipage}[t]{0.5\linewidth}
    \subfigure{
    \begin{minipage}[t]{1.0\linewidth}
    \centering
    \includegraphics[width=\linewidth]{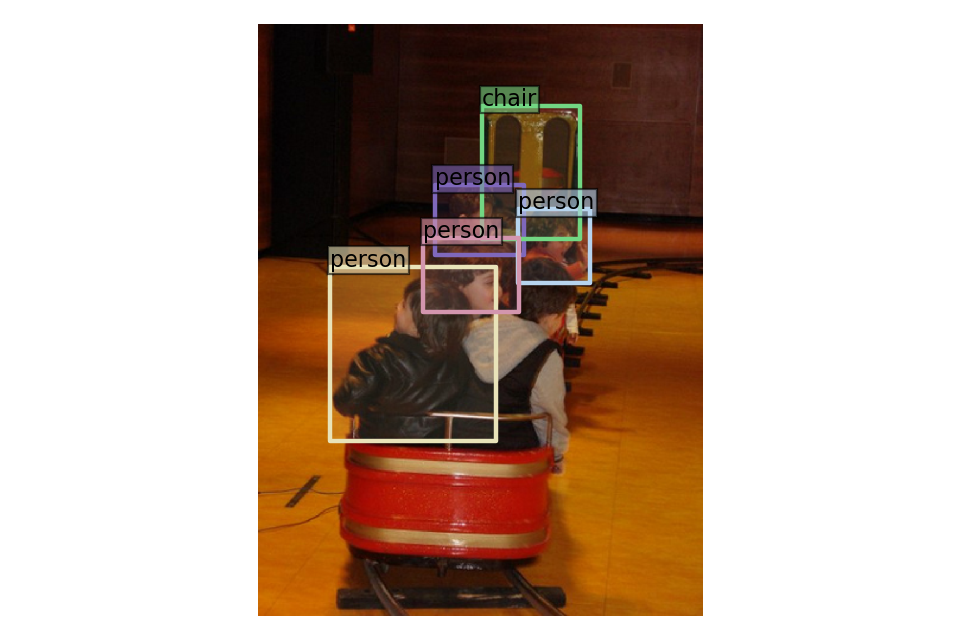}
    \end{minipage}
    }
    \subfigure{
    \begin{minipage}[t]{1.0\linewidth}
    \centering
    \includegraphics[width=\linewidth]{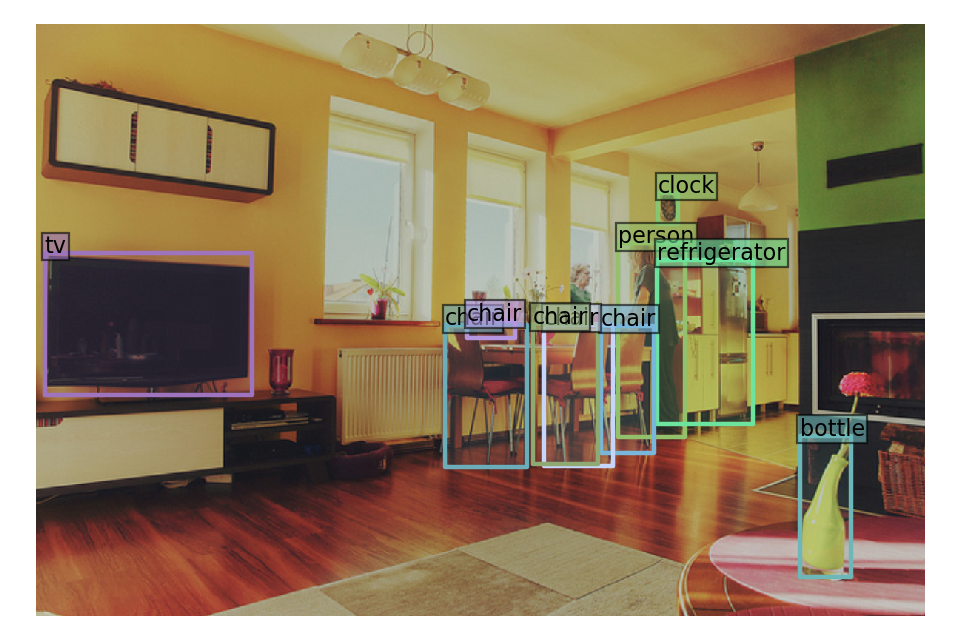}
    \end{minipage}
    }
    \subfigure{
    \begin{minipage}[t]{1.0\linewidth}
    \centering
    \includegraphics[width=\linewidth]{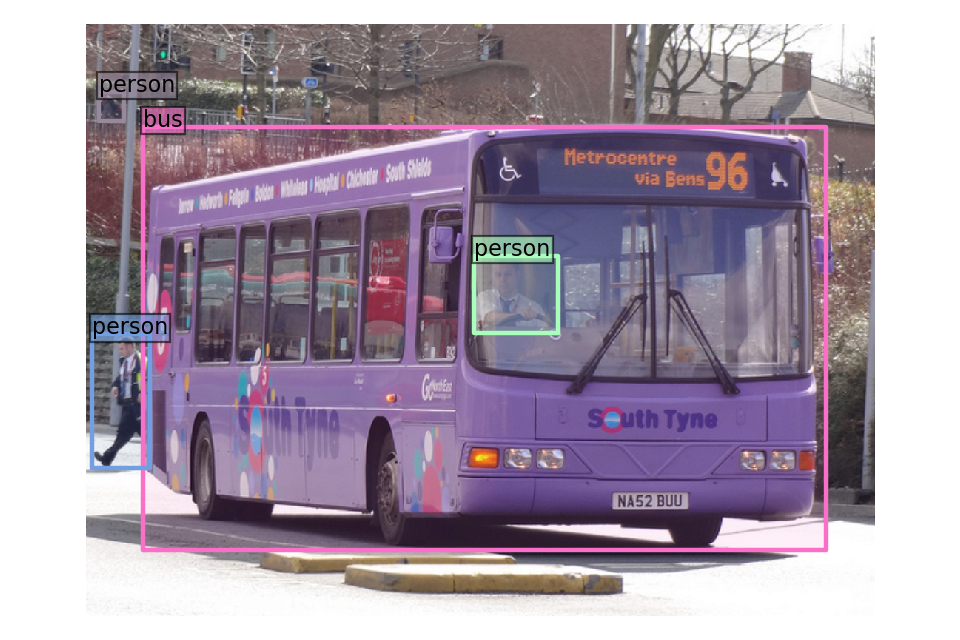}
    \end{minipage}
    }
\end{minipage}
}}
\centering
\caption{Comparisons of outputs of MDETR and our DQ-DETR on COCO detection. Images on the left column are predicted by MDETR, while outputs of DQ-DETR are presented in the right column. We plot predicted bounding boxes whose scores are larger than $0.95$.}
\label{fig:coco_vis}
\end{figure*}

We compare the results on COCO detection in Fig. \ref{fig:coco_vis}.

\section{Limitations} 

Although the remarkable performance compared with MDETR, our DQ-DETR is still inferior to the best DETR-like models on object detection. One main reason for this result is that our model require an understanding of both image and text inputs, yield a more challenging task than object detection. \\
\section{Data usage \& Social impacts.} 

All the datasets we used in our experiments are publicly available and we use the data for academic purposes only. Hence we use the data in accordance with the requirements of their licenses. Phrase grounding involves two modalities, vision, and language. Therefore, it may suffer from common problems in the two fields, like adversarial examples and biased data.

\end{document}